\documentclass[11pt]{article}

\usepackage[]{acl}

\usepackage{times}
\usepackage{latexsym}

\usepackage[T1]{fontenc}
\usepackage[utf8]{inputenc}

\usepackage{microtype}

\usepackage{inconsolata}

\usepackage{graphicx}
\usepackage{amsmath}
\usepackage{booktabs}
\usepackage{subcaption}
\usepackage{multirow}
\usepackage{array}
 \usepackage{amssymb}
\usepackage{tabularx}
\usepackage{float}
\usepackage{tcolorbox}

\usepackage{xcolor}

\title{Domain-Specific Jargon in Large Language Models: A Comparative Analysis between General-Purpose and Specialist Models}

\author{
 \textbf{Darin Keng\textsuperscript{1, 2}},
 \textbf{Zhewei Sun\textsuperscript{2}}
\\
\\
 \textsuperscript{1}Department of Statistics, University of Chicago, Chicago, Illinois\\
 \textsuperscript{2}Toyota Technological Institute at Chicago, Chicago, Illinois\\
 \ttfamily dkeng@uchicago.edu, zsun@ttic.edu
}

\begin{document}
\maketitle
\begin{abstract}
Large Language Models (LLMs) have shown remarkable proficiency on general-purpose tasks, yet their performance often degrades in highly-specialized technical domains. Moreover, little is known about how parametric knowledge of domain-specific terms is encoded within these models. We address this gap by contributing two novel medical jargon evaluation benchmarks and evaluate a general-purpose Llama-3.1 model against a variant fine-tuned on medical-domain data. Surprisingly, the general-purpose model outperforms the medically fine-tuned model on both tasks. Using mechanistic interpretability tools, we find systematic patterns of miscalibration for the medically fine-tuned model.
%we find systematic patterns of over-classification bias on jargon identification and miscalibration on jargon understanding for the medically-finetuned model.
Instead of reorganizing parametric knowledge, the fine-tuned model places greater emphasis on a small subset of model components associated with jargon-favoring predictions. 
% fine-tuning simply amplifies the jargon-favoring tendencies of a small set of model components.
%. exhibiting a systematic over-classification bias on jargon identification and miscalibration on jargon understanding. Applying mechanistic interpretability tools, we find that the architectural locus of jargon knowledge is largely shared between the two models. Instead of reorganizing the parametric knowledge, fine-tuning amplifies the jargon-favoring tendencies of a small set of model components. 
%A targeted component reweighting strategy that suppresses these components yields a 4\% accuracy gain on jargon identification for the fine-tuned model, recovering most of the gap to the general-purpose baseline. 
We find that applying component reweighting strategies against the benchmark tasks successfully suppresses these components and closes the gap with the general-purpose baseline. 
%Finally, a subset of components transfers to a materials science jargon task, suggesting they encode a partially domain-agnostic notion of specialized terminology.
We also observe that some jargon-sensitive components transfer knowledge to the same tasks involving materials science jargon, suggesting they encode a partially domain-agnostic notion of specialized terminology.
Our results provide a case study in which a medically fine-tuned checkpoint does not improve jargon comprehension over its general-purpose counterpart, highlighting that domain adaptation should not be assumed to yield better performance on specialized terminology.
% Our results highlight that domain-specific fine-tuning alone is insufficient in both encoding knowledge of jargon into LLMs and achieving good performance in relevant downstream tasks.
\end{abstract}

\section{Introduction}

Domain-specific terminology, or jargon, refers to language that professionals use that often presents significant cognitive barriers to non-expert individuals~\citep{:/content/books/9789027298652}. Jargon can take the form of a specialized term coined for technical use, such as \textit{hydromyelia} in medicine. Alternatively, jargon can also reuse an existing word that takes on an entirely different meaning. For example, the common word \textit{tears} can refer to physical ruptures in muscle tissue.  
In recent years, the natural language processing (NLP) community has seen a growing interest in domain-specific language~\citep{wang2023grammar, yang2023empower, 10.1145/3764579}, yet little attention has been paid to understanding how LLMs internally represent and acquire knowledge about domain-specific jargon.
%Current research has focused heavily on adapting LLMs to highly technical environments. Techniques such as domain-specific fine-tuning~\citep{wu2024pmc,singhal2023large} and retrieval-augmented generation~\citep{cabello2024meg} have been explored to improve performance on tasks like medical question-answering~\citep{singhal2023large, lucas2024reasoning} and scientific literature summarization~\citep{taylor2022galactica}.

%Yet, despite a growing body of work on model interpretability~\citep{alain2016understanding, wang2022interpretability, geva2022transformer}, little attention has been paid to understanding how LLMs internally represent and acquire knowledge about domain-specific jargon.

\begin{figure}[t]
    \centering
    \includegraphics[width=0.45\textwidth]{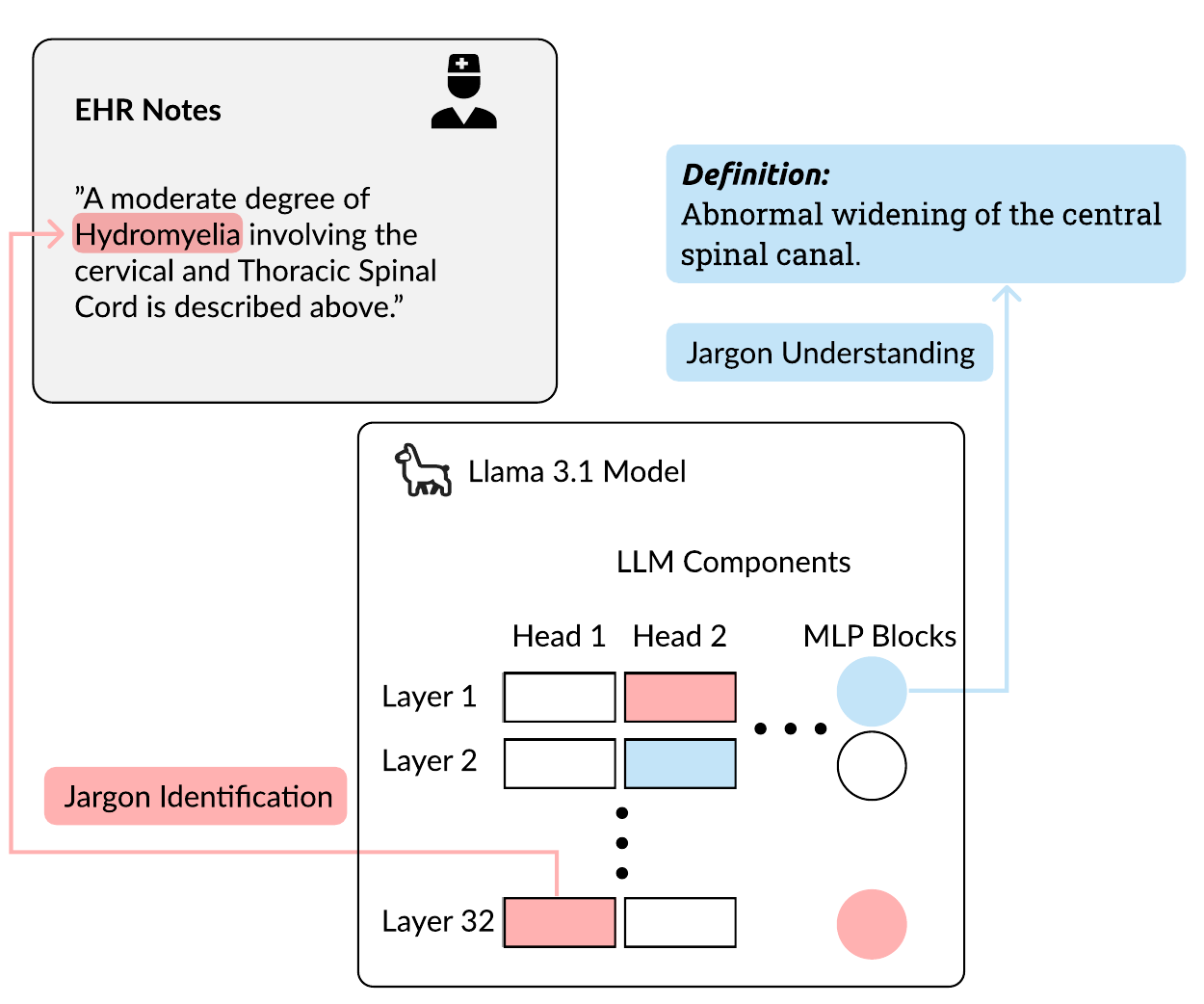}
    \caption{Overview of our framework. Given an EHR note containing a medical term  (\textit{hydromyelia}), Llama-3.1-8B model is evaluated on two tasks: \textcolor{blue}{Jargon Understanding} (selecting the correct definition) and \textcolor{red}{Jargon Identification} (classifying whether the term is jargon in context). The model's outputs are then decomposed into per-component contributions from individual attention heads and MLP blocks to localize jargon knowledge.}
    \label{fig:diagram}
\end{figure}

%As LLMs have been more frequently deployed in specialized settings, a fundamental question arises: To what extent can these models process domain-specific jargon? 

Specialized terminology can impede effective communication between experts and the general public ~\citep{vilhena2014finding,ryba2021better,cervetti2015factors,bullock2019jargon}. 
%Consequently, the ability to accurately interpret specialized terminology is essential for effective communication and decision-making, both for human practitioners~\citep{shulman2020effects} and the AI systems designed to assist them~\citep{august2022generating, rakedzon2017automatic}. 
%In high-stakes domains such as medicine, law, and scientific research, such limitations can have significant practical consequences. Consider a clinical AI assistant designed to summarize Electronic Health Records (EHRs) to patients. If the model misinterprets domain-specific medical jargon,  it may inadvertently provide patients with misleading summaries of their health data~\citep{croxford2025evaluating}.
In high-stakes domains such as medicine, the ability to accurately interpret specialized terminology has significant practical consequences. Consider a clinical AI assistant designed to summarize Electronic Health Records (EHRs) for patients. If the model misinterprets domain-specific medical jargon, it may inadvertently provide patients with misleading summaries of their health data~\citep{croxford2025evaluating}. 
Preventing such failure modes requires not only measuring performance, but also understanding why models succeed or fail. 
%Yet, despite a growing body of work on model interpretability~\citep{alain2016understanding, wang2022interpretability, geva2022transformer}, little attention has been paid to understanding how LLMs internally represent and acquire knowledge about domain-specific jargon. Most existing studies evaluate performance directly at the output level, through tasks such as terminology extraction~\citep{kwon2022medjex} or jargon definition generation~\citep{huang2022understanding}, without examining the underlying mechanisms that give rise to these behaviors. 
%Without such analysis, it is difficult to diagnose failure modes or design more effective training strategies.
To address this gap, we investigate the internal representations of domain-specific terminology in the Llama-3.1-8B model family~\citep{grattafiori2024llama3herdmodels}. We construct two complementary benchmarks that probe distinct aspects of jargon comprehension: 1) A jargon understanding (JU) task that evaluates a model's ability to interpret domain-specific meanings of jargon and 2) A jargon identification (JI) task that tests a model's ability to correctly identify jargon usages in text\footnote{Code and data available at: \url{https://github.com/darinkeng/domain-jargon-llm}}. Figure~\ref{fig:diagram} shows examples of these tasks on an EHR note.

Using these benchmarks, we compare the performance of a general-purpose Llama-3.1-8B-Instruct model against Llama-3.1-8B-UltraMedical~\citep{zhang2024ultramedical}, a domain-adapted variant explicitly fine-tuned on medical text. Moving beyond behavioral evaluation, we adopt a mechanistic approach introduced by~\citet{chang-etal-2024-parts} to decompose the model's outputs into individual contributions of its internal components. By evaluating individual components against our benchmark tasks, we trace how parametric knowledge of jargon is encoded within the model.

To preview our results, we find that the medically fine-tuned UltraMedical model shows inferior performance compared to the general-purpose Llama-3.1 model on both benchmarks. Our analyses show that the two checkpoints retain similar patterns of jargon-sensitive components, while UltraMedical places greater emphasis on components associated with jargon-favoring predictions and increased miscalibration~\cite{guo2017calibrationmodernneuralnetworks, jiang-etal-2021-know}.
% Specifically, our analyses show that domain-specific fine-tuning does not significantly alter behaviors of individual components within the model. Instead, fine-tuning reweights jargon-sensitive components within the model that often results in miscalibration~\cite{guo2017calibration, jiang2021can}.

% Building upon these mechanistic insights, we apply a component reweighting strategy that modulates the influence of jargon-sensitive components and demonstrates improvements in model accuracy on downstream jargon identification. Finally, to test whether these components reflect medical knowledge specifically or a more general notion of specialized terminology, we evaluate them on a parallel jargon identification task in materials science.

We make the following contributions in this paper: 1) Two complementary benchmark tasks for evaluating jargon comprehension for LLMs; 2) A comparative evaluation between general-purpose and domain-specific fine-tuned LLMs on their abilities to process jargon across medical and materials science domains%\revision{A case-study comparison of Llama-3.1-8B-Instruct and Llama-3.1-8B-UltraMedical on their abilities to process jargon across medical and materials science domains}
% A comparative evaluation between general-purpose and domain-specific fine-tuned LLMs on their abilities to process jargon across medical and materials science domains
; and 3) Mechanistic interpretation of LLMs against jargon-related tasks to better understand how LLMs encode parametric knowledge about domain-specific terms.
%identify practical insights to improve LLMs' abilities in processing jargon.
% \begin{itemize}
% \item \textbf{Two complementary benchmarks for jargon comprehension}: Jargon Understanding (JU), a multiple-choice definition-selection task, and Jargon Identification (JI), a context-sensitive binary classification task.
% \item \textbf{Comparative evaluation of general-purpose and medically fine-tuned LLMs}: We compare Llama-3.1-8B-Instruct against Llama-3.1-8B-UltraMedical on both benchmarks, finding that medical fine-tuning degrades rather than improves jargon comprehension and characterizing the resulting behavioral biases.
% \item \textbf{Component-level mechanistic analysis}: Using early decoding and ablation, we localize jargon-relevant knowledge within the model and compare its distribution across the two variants.
% \item \textbf{Targeted component reweighting intervention}: We demonstrate that the behavioral bias introduced by fine-tuning can be mitigated without retraining by reweighting a small set of components.
% \item \textbf{Cross-domain generalization test}: We construct a parallel materials science jargon benchmark (Mat-JI) and test whether the identified components encode a medical-specific or domain-general notion of specialized terminology.
% \end{itemize}

\section{Related Work}
\subsection{Processing jargon with LLMs.} 
As LLMs are increasingly deployed in high-stakes environments such as healthcare and scientific research, their ability to process specialized vocabulary has become a key capability.  Domain-specific fine-tuning has emerged as the most common approach for improving this capability \cite{chen2023meditron70bscalingmedicalpretraining, wu2024pmc, zhang2024ultramedical}.
Existing research on jargon in LLMs has largely concentrated on two complementary tasks: jargon detection~\cite{rakedzon2017automatic, 10.1145/3473141.3473225, 10.1145/3488560.3498469, kwon-etal-2022-medjex, guo-etal-2024-personalized} and jargon definition generation~\cite{huang-etal-2022-understanding,august-etal-2022-generating,liu-etal-2021-graphine}.
Beyond these two dominant tasks, \citet{lucy-etal-2023-words} applies word sense induction to quantify scholarly jargon and \citet{papandreou-etal-2025-medical} investigate jargon-aware prompting strategies for LLM-based medical text simplification. 

These efforts have substantially advanced understanding of how LLMs detect jargon and generate accessible explanations, yet little is known about how internal mechanisms of LLM give rise to the parametric knowledge used to process jargon. Instead of designing tasks with the goal of improving downstream accuracy, our proposed JU and JI tasks provide a simplified setup for testing both contextual awareness and semantic understanding that better suits mechanistic analysis of LLMs. Such analyses allow us to trace parametric knowledge of jargon within internal components of an LLM, giving insights into how internal mechanisms give rise to LLMs' abilities in processing jargon.
%the parametric knowledge used to process jargon.
% Existing research on jargon in LLMs has largely concentrated on two complementary tasks: Jargon detection and jargon definition generation. Jargon detection aims to determine whether a given term constitutes specialized terminology unfamiliar to non-experts~\cite{rakedzon2017automatic, lao2021detecting, ke2022unsupervised, kwon2022medjex, guo2024personalized}. These efforts develop dedicated detection systems as their primary goal, with evaluation focused on the accuracy of the detector itself. Our JI task is an instrument for mechanistic analysis of an existing model rather than a system to be optimized. A parallel line of work focuses on jargon definition generation, which seeks to produce lay-accessible explanations for specialized terms~\cite{huang2022understanding,august2022generating,liu2021graphine}.

%These efforts have substantially advanced understanding of how LLMs detect jargon and generate accessible explanations, yet little is known about how internal mechanisms of LLM give rise to the parametric knowledge used to process jargon. Our work takes a step toward addressing this question by tracing parametric knowledge of jargon within internal components of the LLM, complementing existing behavioral evaluations with a mechanistic perspective.

\subsection{Mechanistic interpretability of LLMs.} 
Mechanistic interpretability tools have been developed to complement behavioral evaluation for LLMs. For example, \citet{meng2022locating} use causal tracing to demonstrate that factual associations are localized within specific mid-layer feed-forward modules.
%, providing an influential framework for localizing knowledge within model components. 
\citet{hewitt-liang-2019-designing} introduce control tasks associating word types with random outputs to establish probe selectivity.
%, thereby demonstrating whether a model is genuinely extracting linguistic structure from internal representations or by simply memorizing the task.
Recent work by \citet{marinescu2026medicalinterpretabilityknowledgemaps} utilizes activation patching and layer lesioning to construct internal knowledge maps of medical LLMs, demonstrating that specialized concepts localize in distinct network layers and highlighting the need to apply interpretability techniques directly to domain-specific vocabulary. 

In this paper, we adopt the interpretation framework proposed by \citet{chang-etal-2024-parts}. This method provides an efficient tool for decomposing the output of an LLM into direct contributions of its internal components — Specifically, the Multi-Layer Perceptron (MLP) blocks and individual attention heads.% \revision{We pair this interpretation framework with our jargon benchmarks to characterize differences in jargon processing between the general-purpose and medically fine-tuned Llama-3.1-8B checkpoints.}
We pair the interpretation framework with our jargon benchmark tasks to gain mechanistic insights into how domain-specific finetuning affects processing of jargon in LLMs.

\section{Jargon evaluation benchmark}
\label{sec:section-3}
\subsection{Data sources}
\label{sec:section-3README}
We create our benchmark using README~\cite{yao-etal-2024-readme}, a large dataset of medical jargon entries. To complement our analysis, we also use MatScholar~\cite{song-etal-2023-matsci}, a materials science jargon dataset, to construct a parallel jargon identification task for cross-domain analysis.% We also use the BoolQ~\cite{clark2019boolq} task as a non-jargon control task.
%We draw on three datasets to construct our evaluation benchmarks. README~\cite{yao-etal-2024-readme} provides the medical jargon used for our two primary benchmarks. MatScholar~\cite{song2023matsci} supplies the materials science data used to construct a parallel jargon identification task for cross-domain analysis. BoolQ~\cite{clark2019boolq} serves as a non-jargon reading-comprehension control to distinguish jargon-specific signal from general task competence. 

\paragraph{README.}
We use \texttt{README-exp\_good}, a 113,659 entry subset of README retrieved from the Unified Medical Language System (UMLS; ~\citealp{10.1093/nar/gkh061}). The underlying jargon annotations in README were manually created by domain experts, who read sampled EHR sentences and identified terms considered difficult to comprehend for individuals with no more than a seventh-grade education. \texttt{README-exp\_good} retains these expert-annotated instances whose UMLS-retrieved definitions passed README’s subsequent quality-filtering procedure.

% that has passed \citet{yao-etal-2024-readme}'s internal quality check. 
Each entry consists of four fields: 1) a medical jargon drawn from an EHR note, 2) the surrounding EHR context, 3) a jargon definition retrieved from UMLS, and 4) a patient-accessible lay definition. Our pipeline uses only the term, the EHR context, and the jargon definition. See Appendix~\ref{app:readme-detail} for further details on how the dataset is used.

%Inspection of \texttt{README-exp\_good} revealed that some terms are widely familiar to laypersons, making them poor examples of specialized jargon. 
To ensure the benchmark contains genuinely specialized terminology, we further filter \texttt{README-exp\_good} using Sentence-BERT (SBERT; \citealp{reimers-2019-sentence-bert}) and Wiktionary~\cite{ylonen22}. Each definition entry in Wiktionary contains tags that indicate the topic or usage context (e.g., medicine, colloquial). We use these tags to check whether a term's medical meaning is too close to its common meaning. If a README entry's jargon definition is sufficiently similar to one of the non-medical Wiktionary definition entry of the corresponding lexical item, we consider the entry a non-jargon entry. See Appendix~\ref{app:wiktionary-tags} for more details. After de-duplication and filtering, we obtain a jargon dataset $\mathcal{J}$ with 5977 entries and a non-jargon dataset $\mathcal{N}$ with 2095 entries. We partition $\mathcal{J}$  into train, test splits, which are used to construct the downstream benchmark tasks described below. Full data statistics are provided in Appendix~\ref{app:data-stats}.

% Let $\mathcal{T}_\text{tech}$ denote the set of Wiktionary tags that indicate specialized technical usage\footnote{See Appendix~\ref{app:wiktionary-tags} for the full list of technical tags}. For each term $w$ in \texttt{README-exp\_good}, we retrieve its set of Wiktionary definitions $D_{wik}$. We define the subset of conventional (non-medical) definitions as:
%     \begin{equation}
%     D_{conv} = \{d \in D_{wik} \mid \text{tags}(d) \cap \mathcal{T}_\text{tech} = \emptyset\}\;
%     \end{equation}
%      We retain $w$ as jargon if its README medical definition $d_w$ is sufficiently distant from all conventional senses:
% \begin{multline}
%     \mathcal{J} = \{ w \in \mathcal{T} : \phi(w) < \tau \}, \\
%  \text{where } \phi(w) = \max_{d \in D_{\text{conv}}} \mathrm{sim}(f_\theta(d_w), f_\theta(d))
% \end{multline}

%     \noindent where $f_{\theta}$ represents the SBERT embedding, $\text{sim}(\cdot)$ is the cosine similarity function, and $\tau = 0.5$ chosen by manual inspection. Terms whose medical definition differs sufficiently from their non-medical senses (similarity $< \tau$) are retained as jargon. Terms that fail this criterion are reassigned to the non-jargon set. After filtering, we obtain a jargon dataset $\mathcal{J}$ with 5977 entries and a non-jargon dataset $\mathcal{N}$ with 2095 entries. We partition J into train, test, and dev splits, which are used to construct the downstream benchmark tasks described below. Full data statistics are provided in Appendix~\ref{app:data-stats}.

\paragraph{MatScholar.}
MatScholar is a named entity recognition (NER) corpus of materials science paper abstracts. Each entry consists of a tokenized sentence paired with a sequence of BIO tags.
%, where B- and I- tags mark the beginning and continuation of domain entity spans, and O marks tokens outside any entity.  
Because MatScholar provides only span annotations but no dictionary definitions, we cannot construct a JU benchmark using this dataset. We therefore only use MatScholar to construct a JI benchmark for the purpose of testing cross-domain generalization of model components. 
%we cannot apply the SBERT-based filtering used to construct the medical benchmarks, nor can we construct a JU-style task that requires definition candidates. We therefore use MatScholar to construct only a parallel JI benchmark. 

% \paragraph{BoolQ.} 
% BoolQ is a binary reading-comprehension benchmark in which each item pairs a short passage with a yes/no question. We use it as a non-jargon control: solving BoolQ requires general language understanding rather than recognition of specialized terminology.

\subsection{Jargon Understanding (JU) task}
\label{sec:ju-task}
% We first construct a jargon understanding (JU) benchmark that tests models' abilities in correctly interpreting a jargon's intended meaning with respect to its usage context.
%We first construct a jargon understanding (JU) benchmark that tests models' semantic understanding of jargon terms.
We first develop a jargon understanding (JU) benchmark to assess models’ semantic comprehension of jargon terms.
We set up the task as multiple-choice question (MCQ) prompts, a prevailing approach for evaluating LLMs \cite{hendrycks2021measuring, zhang2024multiplechoicequestionsefficientrobust} that restricts the output space to a predefined set of options and thus avoids the scoring complexity of free-form generation. For every MCQ, the model is presented with a specific jargon term from $\mathcal{J}$ alongside an EHR note as context. The model must then select the correct definition from five candidate options. %Including the EHR context allows us to test how the model handles polysemy. Many medical terms take on different meanings depending on the clinical scenario. For example, the term \textit{discharge} could mean the release of a patient from the hospital, or it could refer to a bodily exudate (e.g., purulent discharge). The EHR context forces the model to select a definition appropriate to the clinical scenario rather than defaulting to the term's most common sense. 
An example prompt template for JU can be found in Appendix~\ref{app:ju-example}.

To create distractors, we rank all non-target definitions in the training pool by SBERT cosine similarity to the target definition $d_w$ in descending order. To prevent near-duplicates of the target from being selected as distractors, we exclude any candidate whose similarity to $d_w$ exceeds an upper threshold $\tau_{\max} = 0.9$. The distractor set for jargon term $w$ is then defined as the top four ranked candidates that fall below this threshold.
The target and the four distractors are then assigned to labels A-E, with the correct position distributed uniformly across items to prevent position bias. We construct the JU benchmark for both Train and Test splits. Distractors for both splits are drawn exclusively from the Train-set definitions to prevent test-time leakage. In total, we construct 4,183 training and 1,495 testing JU examples. 
% To create challenging distractors, we employed SBERT to dynamically generate hard negatives. We define $\mathcal{D}_w$ as the candidate pool of non-target definitions (i.e., all definitions in the Train data other than $d_w$), comprising a total of $N = |\mathcal{C}_w|$ available options. To identify challenging distractors, we compute the cosine similarity between the SBERT embedding of the target definition and each candidate definition. The candidate definitions are ranked in descending order of their semantic similarity to the target definition, formally expressed as:

% \begin{align}
% \mathrm{sim}(f_\theta(d_w), f_\theta(d_{(1)})) 
%   &\geq \cdots \notag \\
%   &\geq \mathrm{sim}(f_\theta(d_w), f_\theta(d_{(k)}))
% \end{align}
% where $d_{(i)}$ denotes the $i$-th ranked definition in the pool. To prevent near-duplicates of the target definition from being selected as distractors, we exclude any candidate whose similarity to $d_w$ exceeds an upper threshold $\tau_{\max} = 0.9$. The distractor set for jargon term $w$ is then defined as the top four ranked candidates that fall below this threshold:
% \begin{multline}
%     \mathcal{D}_{dist}(w) = \{ d_{(1)}, d_{(2)}, d_{(3)}, d_{(4)} \}, \\ \text{where } \text{sim}(f_{\theta}(d_w), f_{\theta}(d_{(i)})) < \tau_{\max}
% \end{multline}
% where $d_{(i)}$ denotes the $i$-th ranked definition in the pool.

%This ensures the distractors are semantically close to the correct answer, forcing the model to discriminate at a fine-grained clinical level rather than relying on broad topical associations.

\subsection{Jargon Identification (JI) task}
We also construct a jargon identification (JI) benchmark to evaluate the models' abilities to situate jargon usages in linguistic contexts. We set up the task as Binary question answering (Yes/No), a standard paradigm for evaluating factual knowledge in LLMs \cite{hendrycks2021measuring, kamalloo-etal-2023-evaluating}.
%, and closed-set formats have been used effectively to probe domain-specific comprehension in biomedical settings~\cite{jin2019pubmedqa, jin2021disease, tsatsaronis2015overview}. 
Specifically, given a jargon term and its surrounding context, the model must determine whether the term functions as a jargon in that context. We prompt the model to define jargon as terminology that a layperson is unlikely to understand. The evaluation dataset is constructed using a balanced binary sampling strategy. Positive instances are sampled from the jargon set $\mathcal{J}$. Negative instances are sampled from a non-jargon set $\mathcal{N}$. An example prompt of the JI task can be found in Appendix~\ref{app:ji-example}.
We allocate 1,495 negatives to the test split to maintain a 1:1 class ratio, since primary evaluation occurs on the Test set. The remaining 600 negatives are allocated to the train split for component reweighting, along with 600 randomly sampled positives from $\mathcal{J}$. In total, the JI benchmark contains 1,200 training examples and 2,990 test examples.

% In the cases involving reused terms, the JI task also tests context sensitivity: A term may invoke one of its conventional senses in one setting but takes on a specialized jargon meaning in another (e.g., \textit{stable} in a general report versus a clinical assessment). This evaluates whether the model interprets the term in isolation or as a function of its surrounding clinical context.

%Because the non-jargon set $\mathcal{N}$ contains only 2,095 terms, negatives for JI are subject to a fixed budget. 

\subsection{Cross-domain benchmark construction}
To test whether jargon-relevant components generalize beyond the medical domain, we construct a similar JI benchmark using MatScholar.
% and denote the task as Mat-JI.
Tokens or token spans tagged B- or I- are taken as the positive class. For the negative class, we use LLM-as-a-judge procedure to avoid yielding trivial negatives. We score O-tagged tokens on a 1–5 scale of perceived technical complexity, where 1 corresponds to basic English (e.g., the, however) and 5 to deceptive distractors that resemble specialized vocabulary but are not materials science entities (e.g., stochastic, orthogonal). To minimize label noise that could arise from sampling exclusively at the upper end of the scale, we draw negatives uniformly from scores 3–5. The full rubric and prompt are provided in Appendix~\ref{app:mat-ji-app}.
We also use BoolQ~\cite{clark-etal-2019-boolq}, a reading-comprehension benchmark with yes/no questions, as a non-jargon control task. Solving BoolQ requires general language understanding rather than recognition of specialized terminology.
We reformat BoolQ to match JI's A/B prompt structure.

\section{Methodology}
\label{sec:methodology}
%\paragraph{Component decomposition}
Our mechanistic analyses rely on the component decomposition and reweighting framework proposed by~\citet{chang-etal-2024-parts}, which expresses an LLM's output logits as a linear sum of contributions from individual attention heads and MLP blocks. Concretely, for a transformer with L layers, the final hidden state is the sum of the per-layer attention and MLP outputs that each component writes to the residual stream. The output logits can be rewritten as
\begin{equation}
\text{logits} =  \sum_{j}^{N} w_j (U_Y \cdot C_j )
\end{equation}
Where $C_j$ is the post-layernorm activation of component $j$, $U_Y$ is the output embedding matrix restricted to a task's label words, and $w_j$ are the weights assigned to each component (1 by default). The term $U_Y \cdot C_j$ is interpreted as component $j$'s direct contribution to the output logits, enabling per-component accuracy measurements and per-component reweighting. We note that this captures only the direct contribution of each component to the output, not the indirect influence it exerts by modifying the residual stream read by subsequent layers.
For both JU and JI, we score each choice option by reading the model's logit at the final input position over the option-letter tokens (A–E for JU, A/B for JI) and selecting the argmax.
%No sampling or free-form generation is performed, so results are deterministic given the input.

Under this framework, each component in the model can be reweighted by adjusting the learnable weight parameters $w_j$. We follow \citet{chang-etal-2024-parts} and optimize these weights on a training set by minimizing cross-entropy loss with L1 regularization. All other model parameters remain frozen throughout the reweighting process. Because the decomposition is linear, reweighting can be applied directly without retraining the model, making the procedure efficient at scale.

\begin{table}[t]
\small
\centering
\resizebox{\columnwidth}{!}{%
\begin{tabular}{l r r} 
\textbf{Model} & \textbf{JU} & \textbf{JI} \\ 
\midrule
Llama-3.1-8B-Instruct & \textbf{76.1} [73.8, 78.2]& \textbf{58.8} [57.0, 60.6]\\
Llama-3.1-8B-UltraMedical & 74.1 [71.8, 76.3]& 53.1 [51.3, 54.9]\\
\end{tabular}%
}
\caption{Prediction accuracy on both jargon evaluation tasks, with 95\% Wilson confidence intervals in parentheses. Bold indicates the best score.}
\label{tab:eval_task_comparison}
\end{table}

\section{Results}
Unless otherwise noted, all results in this section are reported on the held-out test splits described in Section \ref{sec:section-3} (1,495 items for JU; 2,990 items for JI).

\subsection{Comparative analysis}% of medical and general-purpose LLMs}
\label{sec_comp_analysis}

%Before investigating internal mechanisms, 
We first establish whether the medically fine-tuned UltraMedical exhibits improved performance
% medical fine-tuning produces the behavioral improvement 
on jargon tasks that domain adaptation is typically assumed to deliver. Table~\ref{tab:eval_task_comparison} shows the benchmark task performance of both the general-purpose Llama-3.1-8B-Instruct model and the domain-adapted Llama-3.1-8B-UltraMedical model. Across both tasks, the base Llama-3.1 Instruct model outperforms the domain-adapted UltraMedical model, with the gap being most pronounced on JI. This counterintuitive result motivates the analyses that follow: We examine model calibration on both tasks and error structure on JI to characterize the behavioral differences between the two models.
% how medical fine-tuning shapes model behavior.

\begin{figure}[t]
    \begin{subfigure}[b]{0.5\textwidth}
        \centering
        \caption{Jargon understanding (JU)}
        \includegraphics[width=\textwidth]{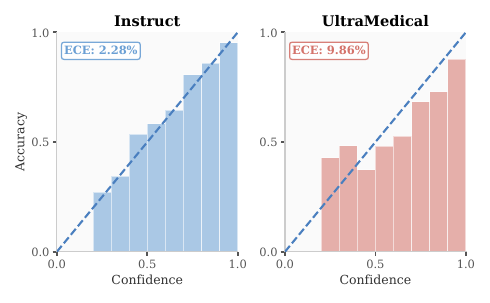}
        
    \end{subfigure}
    \begin{subfigure}[b]{0.5\textwidth}
        \centering
        \caption{Jargon identification (JI)}
        \includegraphics[width=\textwidth]{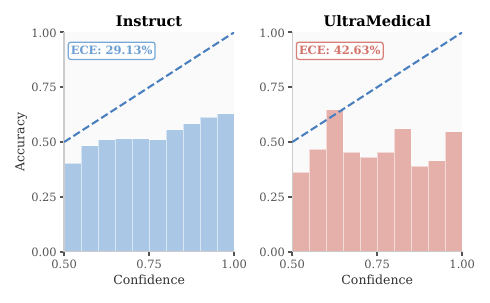}
        
    \end{subfigure}
    \caption{Model calibration on both tasks. The dashed diagonal lines represent perfect calibration.}
    \label{fig:calibrate_curve}
\end{figure}

%The UltraMedical model (ECE = 9.86\%) falls below the diagonal, while the base model (ECE = 2.28\%) closely tracks it.

% \begin{figure}[t]
%     \centering
%     \includegraphics[width=0.5\textwidth]{Images/calibration_curves_onecol_JI.pdf}
%     \caption{Jargon Identification (JI) Calibration Results. Both models are substantially overconfident, with UltraMedical (ECE = 42.63\%) showing worse calibration than the base Instruct model (ECE = 29.13\%)}
%     \label{fig:calibrate_curve}
% \end{figure}

\begin{figure*}[t]
    \centering
    \includegraphics[width=0.95\textwidth]{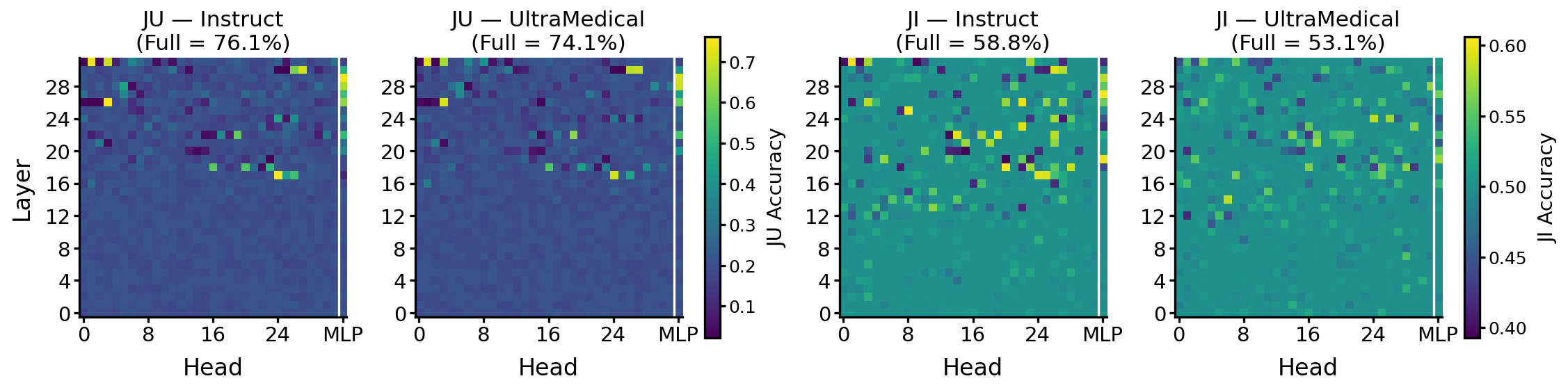}
    \caption{Per-component accuracy for Llama-3.1-8B Instruct and UltraMedical on JU (left) and JI (right). }
    \label{fig:heatmap_all}
\end{figure*}

\paragraph{Calibration analysis.}
% \begin{figure}[t]
%     \centering
%     \includegraphics[width=0.6\textwidth]{Images/calibration_curves.png}
%     \caption{Jargon Understanding (JU) Calibration Results. The dashed diagonal represents perfect calibration. The UltraMedical model (ECE = 9.86\%) falls below the diagonal, while the base model (ECE = 2.28\%) closely tracks it.}
%     \label{fig:calibrate_curve}
% \end{figure}
A well-calibrated model assigns confidence scores that match its empirical accuracy: Predictions made with 80\% confidence should be correct 80\% of the time~\cite{guo2017calibrationmodernneuralnetworks, jiang-etal-2021-know}. Miscalibration is particularly consequential in clinical settings, where downstream decisions depend not only on model outputs but on the reliability of the confidence those outputs carry. 
%Here, a well-calibrated but less accurate model can be considered more useful than a less-calibrated but more accurate model. 
We measure calibration using Expected Calibration Error (ECE; \citealp{naeini15}): The weighted average gap between confidence and accuracy across binned predictions.
Results in Figure \ref{fig:calibrate_curve} show that, on both tasks, the two models differ sharply in calibration quality. The Instruct model tracks the diagonal more closely, while UltraMedical's accuracy is consistently lower in higher-confidence bins, which is the standard signature of overconfidence.
%This suggests that the base model is appropriately uncertain when it lacks the knowledge to answer correctly, whereas the medical model is poorly calibrated in the face of uncertainty. 
In a clinical setting, where downstream decisions depend on the reliability of model confidence, this form of miscalibration %\revision{represents a potential drawback of the UltraMedical checkpoint studied here}
represents a potential drawback of domain adaptation: The model's output becomes harder to trust precisely in the high-confidence region where users would be more inclined to believe the model.

% \begin{figure*}[t]
%     \centering
%     \includegraphics[width=0.95\textwidth]{Images/heatmaps_1x4.png}
%     \caption{Per-component accuracy for Llama-3.1-8B-Instruct and Llama-3.1-8B-UltraMedical on JU (left two panels) and JI (right two panels). }
%     \label{fig:heatmap_all}
% \end{figure*}

\paragraph{Error analysis.}
We perform a qualitative error analysis on the JI results to characterize common patterns in the models' errors.
%Here, we aim to identify systematic biases introduced by domain-specific fine-tuning that accuracy alone obscures. 
We focus on disagreement cases: Items where one model answers correctly and the other does not. 
%For JI, we turn to qualitative error analysis. Where calibration captures the reliability of a model's confidence over a clean ground truth, error analysis is better suited to characterizing the direction of a model's mistakes. By examining which terms each model misclassifies, and in which direction, we can identify systematic biases introduced by domain fine-tuning that aggregate accuracy alone obscures. We focus on disagreement cases: items where one model answers correctly and the other does not.
Table~\ref{tab:ji-disagreement} shows clear distinction between the models' errors. Because the two models are evaluated on the same test examples, we use a two-sided exact McNemar test to assess whether the observed difference in paired predictions is statistically significant. Instruct correctly predicts 368 examples that UltraMedical misses, whereas UltraMedical correctly predicts 197 examples that Instruct misses; this asymmetry is statistically significant (\(p < 0.001\)). See more detail about McNemar test in Appendix~\ref{app: stat_detail}.
%The two models' errors on JI are nearly mirror images (Table~\ref{tab:ji-disagreement}). 
% UltraMedical's errors are almost entirely false positives (367 of 368 disagreement cases), while Instruct's are almost entirely false negatives (196 of 197). 
Relative to Instruct, UltraMedical exhibits a decision boundary shifted toward over-classification:
% Medical fine-tuning tends to shift the decision boundary toward over-classification: 
Terms appearing in clinical contexts are systematically treated as specialized. The Instruct model exhibits bias in the reverse direction, occasionally missing context-dependent polysemy in which everyday words take on specialized clinical meanings. Full examples from each category of errors are provided in Appendix~\ref{app:ultramed-error-example} and~\ref{app:instruct-error-example}.

\begin{table}[t]
\centering
\small
\setlength{\tabcolsep}{1pt} 
\label{tab:ji-disagreement}
\begin{tabular}{lrr}
 & \textbf{Instruct} & \textbf{UltraMedical} \\
\midrule
Total Errors & 197 & 368 \\
False Positives & 1  & 367  \\
False Negatives & 196  & 1  \\
Mean Confidence Margin & 1.64 & 3.04 \\
\end{tabular}
\caption{JI Disagreement Error Summary. Each column counts cases where the named model is wrong while the other model is correct. The difference in discordant predictions is significant under a two-sided exact McNemar test (\(p<0.001\)). Absolute Mean Confidence Margin is calculated by averaging the Margin $= | \text{logit}(A) - \text{logit}(B) |$, where $A$ = Jargon and $B$ = Non-Jargon.}
\label{tab:ji-disagreement}
\end{table}
%, restricted to items where the two models disagree

%This directional asymmetry is the central behavioral finding of our comparative analysis: the two models do not simply differ in accuracy, but in the kind of error each is prone to make. 

\begin{figure*}[t]
    \centering
    \includegraphics[width=0.8\textwidth]{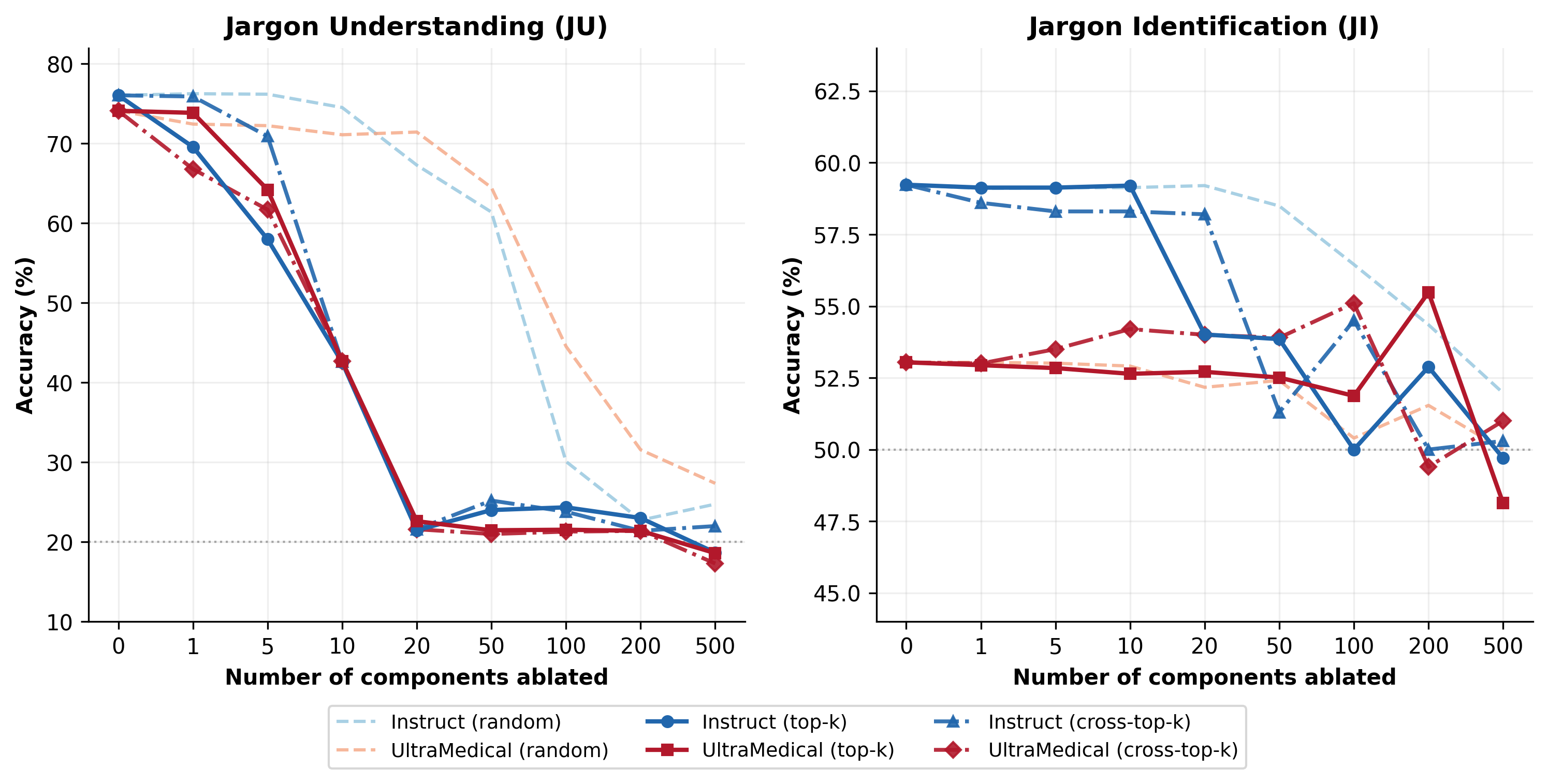}
    \caption{Cumulative ablation results on JU (left) and JI (right). Horizontal gray lines mark chance performance (20\% on JU, 50\% on JI).}
    \label{fig:ablate_cumulative}
\end{figure*}

%Solid lines (top-k) ablate the top-k components ranked by each model's own decomposition accuracy; dash-dot lines (cross-top-k) ablate the top-k components ranked by the other model's decomposition, applied to the current model; dashed lines show the random-ablation control.
%Targeted ablation on JU causes a steep collapse toward chance-level performance by $k$=20, while random ablation degrades accuracy far more gradually. JI shows a much flatter response to both targeted and random ablation, consistent with its more distributed component-level accuracy.

\subsection{Localization of domain-specific knowledge}
\label{sec:comp-analysis}
%The behavioral analysis establishes that UltraMedical is more overconfident than Instruct on our jargon benchmarks.
The behavioral analysis establishes that medical fine-tuning of Llama-3.1-8B-Instruct shifts model behavior toward overconfidence. 
We next investigate whether this shift reflects a reorganization of parametric knowledge encoded within the model.
%, or instead a reweighting of contributions from a shared set of components. 
To examine this, we apply the method described in Section~\ref{sec:methodology} to decompose each model's output into per-component contributions and identify components that carry jargon-relevant signal in each model.

Figure \ref{fig:heatmap_all} shows per-component accuracy for both models on both tasks. For JU, high-performing components are concentrated in the upper layers (17–31) in both Instruct and UltraMedical, with the lower layers contributing near-zero individual accuracy. A small set of components stands out in both models — most notably L17H24, L26H3, MLP-29, and L31H1, which are the top four components for both Instruct and UltraMedical (differing only in rank order). The JI heatmaps exhibit a markedly more sparse distribution, indicating that contextual understanding of jargon is more spread out within the models. We provide the full list of top components in Appendix~\ref{app:top-performing-comp}.

%Most components fall within a narrow 30–50\% accuracy band, with no individual component approaching full-model accuracy. Full top component tables are provided in Appendix~\ref{app:top-performing-comp}.

%The JI heatmaps exhibit a markedly flatter distribution: Most components fall within a narrow 40–60\% accuracy band, with no individual component approaching full-model accuracy.

%\paragraph{Weighted Kendall's $\tau$. } 
\paragraph{Ranked correlations.}
To quantify how similarly the two models rank their components, we compute weighted Kendall's $\tau$~\cite{vigna2015weighted} between the per-component accuracy of Instruct and UltraMedical. Unlike standard Kendall's $\tau$, which weights all rank disagreements equally, the weighted variant places greater weight on disagreements near the top of the ranking, making it well-suited to our setting where the highest-accuracy components are of primary interest. See Appendix~\ref{app:wkendall} for a full specification of the metric. On JU, the two models agree closely at the top ($\tau$ = 0.72, 95\% CI: [0.69, 0.76]). On JI, agreement is slightly weaker ($\tau$ = 0.59, 95\% CI: [0.47, 0.65]) but still substantial. We estimate 95\% confidence intervals using 10,000 paired bootstrap resamples over components. See more detail in Appendix~\ref{app: stat_detail}.
%, consistent with the sparse per-component accuracy distribution and the absence of a small set of dominant components.
Together, these patterns suggest that domain adaptation does not reorganize where jargon knowledge resides. The architectural locus is largely shared between the base and fine-tuned models, with fine-tuning modulating component contributions rather than relocating them\footnote{Following~\citet{chang-etal-2024-parts}, our decomposition measures each component’s direct contribution to the output logits. It does not capture indirect effects in which a component modifies the residual stream and thereby influences downstream components. Thus, our conclusion that the two checkpoints primarily differ through reweighting rather than reorganization applies only to direct component contributions and does not rule out reorganization through indirect pathways.}.

\paragraph{Ablation studies.}
To verify that the components identified above are causally responsible for task performance, we perform a cumulative ablation study in which the top-$k$ components, ranked by individual accuracy, are removed simultaneously from the model. Following~\citet{NEURIPS2019_2c601ad9}, we ablate attention heads and MLP blocks by zeroing their outputs before they are written to the residual stream, removing their direct contributions while leaving all learned weights intact (see Appendix~\ref{app:ablation} for full details).
Figure~\ref{fig:ablate_cumulative} shows the results. 
%On JU, Instruct accuracy drops from 76.1\% to 58.0\%, 42.5\%, 21.5\% when the top 5, 10, 20 components were ablated respectively. 
On JU, both models' accuracy decreases substantially quicker when ablating the top-$k$ components, reaching chance performance by $k=$ 20, whereas performance erodes more gradually when random components are being ablated.
%UltraMedical follows a near-identical trajectory, reaching chance performance around the same threshold. A random-ablation control degrades performance far more gradually. Removing 10 random components costs only 1.5\%, confirming that the targeted ablation is indeed removing key components. 
On JI, by contrast, both targeted and random ablation produce only shallow declines that track one another closely.
The contrast between JU and JI ablation profiles suggests that JU relies on a small set of specialized components whereas JI relies on aggregate contributions from multiple components.
We also cross-ablate each model with top-$k$ components identified by the other model variant. Under both tasks, we find that cross-ablating top-$k$ components closely tracks the conditions where components identified by the corresponding models are ablated, %providing further evidence that the two models rely on largely overlapping components for these tasks, consistent with differences in reliance on shared components rather than relocation of encoded knowledge.
giving further evidence that domain-specific finetuning simply modulates the encoded knowledge instead of reorganizing it. 
Interestingly, when cross-ablating the UltraMedical model, performance on JI increases notably. We postulate that the fine-tuned UltraMedical model relies on components sensitive to jargon semantics to appropriate jargon terms as conventional terms.
%these jargon-related components that contributes to the model's overconfidence.
%consistent with the flatter per-component accuracy distribution: JI performance relies on aggregate contributions from many components rather than a few specialized ones. 

% Together, the results suggests that
% medical fine-tuning does not substantially reorganize which components are sensitive to
% jargon-related knowledge. Rather, the architectural locus of this knowledge appears to be
% largely determined during pretraining, with fine-tuning primarily modulating component
% contributions rather than shifting their distribution across layers.
%Together, these patterns suggest that domain adaptation does not reorganize where jargon knowledge resides. The architectural locus is largely shared between the base and fine-tuned models, with fine-tuning modulating component contributions rather than relocating them.
%The contrast between JU and JI ablation profiles is itself informative: jargon understanding is supported by a small set of specialized components whose removal collapses performance, while jargon identification draws on broad architectural participation.

\subsection{Component reweighting}
\label{sec: comp-rw}
So far, our results suggest that UltraMedical
% domain-specific finetuning 
simply modulates existing jargon-sensitive components within the model. We now investigate whether UltraMedical's component weighting is well aligned with performance on our downstream jargon tasks.
% the modulation learned by domain-specific finetuning is optimal w.r.t the downstream tasks. 
We follow the procedure described in Section~\ref{sec:methodology} to learn component weights. See Appendix~\ref{app:expsetupcomponent} for the detailed training procedure.

%\paragraph{Experimental Setup.} We optimize the component weights $w \in \mathbb{R}^{1056}$ (1,024 attention heads + 32 MLP) separately for each task and each model. Weights are trained for 1000 epochs with SGD (lr = 0.01, $\lambda$ = 0.05). on the JI training set (1,200 examples) and evaluated on the 2,990-item JI test set. The same procedure is applied to JU using its 4,183-item training set and 1,495-item test set. The same procedure is applied to JU. Full JU analysis is deferred to Appendix X.

\paragraph{Results.}
\begin{table}[t]
\centering
\resizebox{\columnwidth}{!}{%
\begin{tabular}{l rr rr}
& \multicolumn{2}{c}{\textbf{JU}} & \multicolumn{2}{c}{\textbf{JI}} \\
\addlinespace[0.1cm]
& Baseline & Reweighted & Baseline & Reweighted \\
\midrule
\addlinespace[0.1cm]
Instruct & 76.1 & 76.9 [74.7, 79.0] & 58.8 & 57.8 [56.1, 59.6]\\
UltraMedical & 74.1 & 74.2 [71.8, 76.2] & 53.1 & 56.1 [54.3, 57.9]\\
\end{tabular}%
}
\caption{Baseline and reweighted accuracy on JU and JI for both models, with 95\% Wilson confidence intervals in parentheses.}
\label{tab:comp_reweight}
\end{table}

 Table~\ref{tab:comp_reweight} shows the results before and after reweighting. Reweighting yields a 3.0\% accuracy gain on JI for UltraMedical that is significant under a two-sided exact McNemar test (\(p=0.01\)). See more detail in Appendix~\ref{app: stat_detail}. For Instruct on JI, the same procedure produces a 1.0\% decrease in accuracy. Gains on JU are marginal for both models ($\leq$0.2\%). Our results show that reweighting the fine-tuned model on JI results in a substantial performance improvement (3.0\%), showing that the component weighting exhibited by UltraMedical is not optimal for JI and can be partially corrected through post-hoc reweighting. The result cautions against assuming that domain adaptation will necessarily yield component configurations optimized for every domain-specific downstream task.
 
 % indicating that fine-tuning does not necessarily learn modulations of model components that are beneficial toward domain-specific downstream tasks.
 % Our results show that while the component modulation is near-optimal on JU, domain-specific fine-tuning results in a more substantial performance improvement (3.0\%) compared to the base model (-1.6\%) on JI. This indicates that fine-tuning does not necessarily learn modulations of model components that are benefitial toward domain-specific downstream tasks.
%Rather than acting as a generic recalibration, the reweighting procedure selectively benefits the model that exhibits a directional bias and slightly hurts the model that does not.
 %We attribute this to JU performance already being saturated by a few high-accuracy components (see Appendix X). 

 %MLP-29 and L30H27 receive the largest weight reductions in both models, but the suppression is substantially more aggressive in UltraMedical (0.46 and 0.47) than in Instruct (0.54 and 0.60). Notably, these are also among the highest-accuracy components on JU (Section~\ref{sec:comp-analysis}; Appendix X), where they contribute correctly to definition selection. 

\begin{table}[t]
    \centering
    \small % Slightly smaller text makes it look much more professional
    \setlength{\tabcolsep}{4pt} % Tightens the horizontal padding
    \begin{tabularx}{\columnwidth}{>{\raggedright\arraybackslash}X r r}

        \textbf{Question} & \textbf{Orig. M.} & \textbf{Rew. M.} \\ 
        \midrule
        PAST MEDICAL HISTORY: Significant for B-cell lymphoma \dots history of methadone use, hypertension, anxiety, deconditioning, nicotine \textbf{addiction}, depression, \dots & 2.01 & 0.0 \\ 
        \addlinespace[0.1cm]
         Assessment of Current Medications: medication administration orders for isosorbide mononitrate \dots Er \textbf{tablet}, By Mouth, BID, Routine, **DATE[Oct 21 07] 10:55:00**; heparin 25000 Unit(s) 250 mL \dots & 1.98 & 0.0
    \end{tabularx}
    \caption{Individual logit contributions of L30H27 and full-model margins for two non-jargon terms incorrectly classified as jargon by the UltraMedical model. Orig. M. and Rew. M. denote the full-model logit margin (jargon $-$ non-jargon) before and after reweighting.}
    \label{tab:L30H27-examples}
\end{table}

\paragraph{Error analysis.} 
% MLP-29 and L30H27 receive the largest weight reductions in both models, but the suppression is substantially more aggressive in UltraMedical (0.27 and 0.39) than in Instruct (0.54 and 0.60). Notably, these are also among the highest-accuracy components on JU (Section~\ref{sec:comp-analysis}; Appendix~\ref{app:top-performing-comp}), where they contribute correctly to definition selection. 
MLP-29 and L30H27 receive the largest weight reductions in both models, but the suppression is qualitatively different. In UltraMedical, the optimization effectively eliminates both components from the JI prediction: L30H27 receives a weight of approximately 0 and MLP-29 a weight of $10^{-4}$. In Instruct, the same components are only partially suppressed, with weights of 0.39 and 0.27 respectively. Notably, these are also among the highest-accuracy components on JU (Section~\ref{sec:comp-analysis}; Appendix~\ref{app:top-performing-comp}).%, where they contribute correctly to definition selection.

By examining the JI error structure, the UltraMedical gain comes from correcting 270 false positives at the cost of introducing 146 new false negatives, a net shift that directly counteracts the over-confidence bias documented in Section~\ref{sec_comp_analysis}. The components being downweighted most aggressively are the same ones whose individual contributions most strongly favor the jargon label. 
To illustrate how this suppression affects individual predictions, Table~\ref{tab:L30H27-examples} shows two non-jargon terms that UltraMedical originally misclassified as jargon. 
%Recall from Section~\ref{sec:methodology} that the full-model logits are a sum of per-component contributions; reweighting scales each component's contribution by its learned weight $w_j$ before summing. 
For both \textit{addiction} and \textit{tablet}, the baseline full-model margin favors the jargon label by roughly +2.0. After reweighting, L30H27's weight is reduced to 0, zeroing out its contributions and flipping both predictions to the correct non-jargon label. The same component that drives accurate definition selection on JU is also the primary source of over-classification on JI, and the reweighting procedure recovers JI accuracy by selectively dampening its contribution at inference time.

For Instruct, the absence of a directional bias means the optimizer has no systematic error pattern to correct. Instruct's baseline JI errors are roughly balanced between false positives and false negatives, and applying the same suppression to MLP-29 and L30H27 produces a corresponding shift toward false negatives without a compensating reduction in false positives. The 1.0\% degradation is the expected consequence: The intervention removes signal that was contributing to balanced predictions, with no bias for the suppression to cancel out.

This asymmetry suggests that UltraMedical
% medical fine-tuning
amplifies the jargon-favoring tendencies of specific pre-existing components, and that the reweighting gain stems from suppressing this amplified bias rather than from a generic recalibration of the model. Instruct, on the other hand, loses performance due to the suppression of components that were not over-amplified to begin with.
%: The reweighting gain on UltraMedical comes from suppressing this amplified bias; the corresponding loss on Instruct comes from suppressing components that were not over-amplified to begin with. 
%The same procedure, applied to the same components, produces opposite outcomes precisely because the bias it targets is present in one model but absent in the other.

% The fact that the same procedure yields only a marginal gain on the already-balanced Instruct model further supports this interpretation.

\subsection{Cross-domain generalization}
Section~\ref{sec:comp-analysis} identifies a small set of components that disproportionately contribute to jargon comprehension on the medical jargon benchmarks. A natural follow-up question is whether these components encode a domain-agnostic notion of specialized terminology, or whether they are specific to the medical domain. To investigate this, we construct a parallel jargon identification benchmark in the materials science domain (Mat-JI) and compare per-component accuracy across the two domains. We additionally include BoolQ~\citep{clark-etal-2019-boolq}, a binary reading-comprehension task reformatted into the same A/B prompt structure, as a non-jargon control to rule out the possibility that the top components reflect general A/B classification competence rather than jargon comprehension. Details about the construction of Mat-JI are provided in Appendix~\ref{app:mat-ji-app}. 

We apply the component decomposition procedure from Section~\ref{sec:methodology} to compute per-component accuracy on Medical Jargon Identification (Med-JI) and Mat-JI for Llama-3.1-8B-Instruct, then filter for components that rank in the top-50 of both jargon tasks but fall outside the top-50 of BoolQ. This filter isolates 9 components whose strong jargon performance is not explained by general task competence. Table~\ref{tab:Jargon specific components} lists these components.

\begin{table}[t]
\centering
\small
\begin{tabular}{l r r r}
\textbf{Component} & \textbf{Med-JI} & \textbf{Mat-JI} & \textbf{BoolQ} \\
\midrule
L20H10  & 0.562 & 0.769 & 0.505 \\
L18H26  & 0.553 & 0.735 & 0.539 \\
L29H16  & 0.532 & 0.673 & 0.533 \\
L26H4   & 0.537 & 0.607 & 0.448 \\
L19H5   & 0.571 & 0.605 & 0.349 \\
L15H11  & 0.552 & 0.619 & 0.531 \\
L26H27  & 0.570 & 0.590 & 0.510 \\
MLP-19  & 0.598 & 0.595 & 0.532 \\
L21H18  & 0.578 & 0.589 & 0.501 \\
\midrule
Full Accuracy & 0.588 & 0.739 & 0.816 \\
\end{tabular}
\caption{Components in the top-50 of both Med-JI and Mat-JI but outside the top-50 of BoolQ.}
\label{tab:Jargon specific components}
\end{table}

Two components stand out as the strongest candidates for domain-agnostic jargon knowledge. L20H10 and L18H26 achieve near full model accuracy on Mat-JI. Meanwhile, their Med-JI accuracies likewise rank among the highest individual contributions while performing near chance at BoolQ. The remaining seven components in Table~\ref{tab:Jargon specific components} satisfy the filter criterion but exhibit weaker jargon signal. These components are concentrated in the upper-middle layers rather than the final layers that dominate the unfiltered rankings on Med-JI alone, suggesting that jargon recognition may operate on top of a separate, later-stage answer-formatting circuit. Together, these findings indicate that a small subset of upper-middle-layer components carries jargon-relevant knowledge that transfers across two unrelated technical domains, supporting the interpretation that domain-agnostic knowledge about jargon is also encoded within these models.
%jargon knowledge is partly organized along domain-agnostic lines within the model.
These findings motivate testing whether interventions developed on medical jargon, such as the reweighting procedure in Section~\ref{sec: comp-rw}, can transfer to other technical domains without per-domain retraining. The presence of this signal in the general-purpose model and its transfer across two unrelated technical domains suggests that at least some jargon-relevant structure is already present prior to domain-specific fine-tuning.
% This partial domain-agnosticism has an important practical implication: Interventions developed on medical jargon, such as the reweighting procedure in Section~\ref{sec: comp-rw}, may transfer to other technical domains without the need to perform per-domain retraining. 
% The fact that this signal is present in the general-purpose model and transfers across two unrelated technical domains suggests that jargon identification is not a capability acquired through domain-specific fine-tuning, but rather a partially domain-agnostic structure already encoded during pretraining.

\section{Conclusion}
We have presented a mechanistic comparison of jargon comprehension in general-purpose and medically fine-tuned LLMs. Using two novel jargon comprehension benchmarks designed for mechanistic interpretation, we evaluate the Llama-3.1-8B family models on the medical domain as a case study. Our results suggest that while domain-specific fine-tuning is widely assumed to improve performance on specialized text, the medically fine-tuned UltraMedical underperforms its general-purpose counterpart on both jargon benchmarks and shows greater overconfidence in domain-specific lexical items.
% it can in fact degrade comprehension by being overconfident in domain-specific lexical items. 
Component-level analysis reveals that this bias does not stem from a reorganization of internal knowledge, but can instead be attributed to placing excessive emphasis on specific jargon-sensitive components, resulting in a re-modulation of existing knowledge that is not optimal for improving performance on related downstream tasks.
%\revision{We additionally identify components in the general-purpose checkpoint whose jargon-relevant signal transfers between medical and materials-science terminology, suggesting that some aspects of jargon processing are shared across technical domains. Taken together, our findings provide a case study showing that domain-specific fine-tuning might not outperform its general-purpose counterpart on specialized terminology, and emphasize the need for more robust methodologies for processing domain-specific language.}
We also observe evidence that pre-trained models possess domain-agnostic knowledge of jargon, again questioning the necessity for domain-specific fine-tuning. Overall, our results underscore the limitations of domain-specific fine-tuning for improving model performance on downstream tasks involving specialized terminology and emphasize the need for more robust methodologies for processing domain-specific language.

% : A small set of components carries jargon-relevant signal in both models, and fine-tuning amplifies their jargon-favoring tendencies rather than relocating them. 
% A targeted reweighting of these components recovers most of the gap to the general-purpose baseline, and a subset of them further transfers to materials science jargon, suggesting partially domain-agnostic structure. Our findings indicate that domain adaptation can fail not by losing knowledge but by distorting how existing knowledge is used.
\section*{Limitations}
We use an SBERT-based filtering pipeline to construct and filter our benchmarks. We acknowledge that automatic methods may not always yield the optimal results but opted to use such a pipeline to enable scale in our evaluation benchmark. While constructing the dataset, we only retain terms whose medical definitions are sufficiently distant from their conventional meanings as jargon, thus prioritizing precision on the positive class. This is appropriate given that our central claims concern model behavior on confidently specialized terminology. However, a tradeoff does exist and it would be interesting to extend the benchmark by having human annotations with annotator confidence scores that would enable finer-grained analysis of more ambiguous cases of jargon usage.

Due to computational resource constraints, our analysis focuses on the Llama-3.1-8B family alone, comparing one general-purpose model against one medically fine-tuned variant. While such a controlled pairing holds
% isolates the effect of fine-tuning while holding 
architecture and scale fixed, UltraMedical’s training data and fine-tuning procedure introduce potential confounds, preventing us from isolating the causal effect of domain-specific fine-tuning itself. We attempted to train a controlled fine-tuned variant, but obtaining a model of comparable quality proved computationally demanding, and the resulting checkpoints were not sufficiently reliable for experimentation. Future work should evaluate additional model families and scales to perform controlled fine-tuning experiments in which model training is explicitly controlled.

% it would be interesting to test whether our findings extend to other model families and model scales.
We also acknowledge that we have only examined English jargon usages and leave multilingual extensions to future work.

% — extend to other model families (e.g., Gemma, Mistral), other medical fine-tunes (e.g., MedGemma, BioMistral), or larger scales.

\section*{Ethics Statement}
README is distributed under CC-BY-NC 4.0 license for research purposes. MatScholar is released under the MIT License, and we use BoolQ under CC-BY-SA-3.0 license. README's EHR contexts were anonymized by the original dataset authors as part of their release. We acknowledge the potential risk for personal information leakage when using medical domain data, but we perform no further identification of individuals to ensure that no additional risk is introduced.
We also did not collect new data from human subjects.  README is distributed for research purposes; our derived benchmarks inherit the same access conditions and are intended only for research on model behavior, not for clinical deployment. %We acknowledge that our analysis covers one model pair, and the directional biases we identify may not transfer to other medical LLMs without per-model verification. 

We used AI assistants to expedite the coding process. All code snippets produced by AI assistants were verified by the authors before being used. For writing, we only used AI assistants for light copy-editing and phrasing suggestions.

\section*{Acknowledgments}
We thank the anonymous ARR reviewers and chairs for their constructive comments and suggestions. We are grateful to Aaron Schein and Mei Wang for valuable feedback throughout the work. We also thank I-An Chang and Yu-Cheng Chang for thoughtful feedback and suggestions on earlier drafts of this paper.

\bibliography{custom}

@book{:/content/books/9789027298652,
   author = "Cabré, Teresa",
   editor = "Sager, Juan C.",pages = "",
   title = "Terminology",
   publisher = "John Benjamins",
   year = "1999",
   url = "https://www.jbe-platform.com/content/books/9789027298652",
}

@article {vilhena2014finding,
author = {Daril A. Vilhena and Jacob G. Foster and Martin Rosvall and Jevin D. West and James Evans and Carl T. Bergstrom },
title = {Finding Cultural Holes: How Structure and Culture Diverge in Networks of Scholarly Communication},
journal = {Sociological Science},
volume = {1},
number = {15},
issn = {2330-6696},
url = {http://dx.doi.org/10.15195/v1.a15},
doi = {10.15195/v1.a15},
pages = {221--23},
year = {2014},
}

@article{ryba2021better,
  title={Better writing in scientific publications builds reader confidence and understanding},
  author={Ryba, Ren and Doubleday, Zo{\"e} A and Dry, Matthew J and Semmler, Carolyn and Connell, Sean D},
  journal={Frontiers in psychology},
  volume={12},
  pages={714321},
  year={2021},
  publisher={Frontiers Media SA},
  doi = {10.3389/fpsyg.2021.714321}
}

@article{cervetti2015factors,
author = {Cervetti, Gina and Hiebert, Elfrieda and Pearson, P. and Mcclung, Nicola},
year = {2015},
month = {11},
pages = {153-185},
title = {Factors That Influence the Difficulty of Science Words},
volume = {47},
journal = {Journal of Literacy Research},
doi = {10.1177/1086296X15615363}
}

@article{bullock2019jargon,
author = {Bullock, Olivia and Colon Amill, Daniel and Shulman, Hillary and Dixon, Graham},
year = {2019},
month = {07},
pages = {845-853},
title = {Jargon as a barrier to effective science communication: Evidence from metacognition},
volume = {28},
journal = {Public Understanding of Science},
doi = {10.1177/0963662519865687}
}

@inproceedings{august-etal-2022-generating,
    title = "Generating Scientific Definitions with Controllable Complexity",
    author = "August, Tal  and
      Reinecke, Katharina  and
      Smith, Noah A.",
    editor = "Muresan, Smaranda  and
      Nakov, Preslav  and
      Villavicencio, Aline",
    booktitle = "Proceedings of the 60th Annual Meeting of the Association for Computational Linguistics (Volume 1: Long Papers)",
    month = may,
    year = "2022",
    address = "Dublin, Ireland",
    publisher = "Association for Computational Linguistics",
    url = "https://aclanthology.org/2022.acl-long.569/",
    doi = "10.18653/v1/2022.acl-long.569",
    pages = "8298--8317"
}

@article{rakedzon2017automatic,
  title={Automatic jargon identifier for scientists engaging with the public and science communication educators},
  author={Rakedzon, Tzipora and Segev, Elad and Chapnik, Noam and Yosef, Roy and Baram-Tsabari, Ayelet},
  journal={PLOS ONE},
  volume={12},
  number={8},
  pages={e0181742},
  year={2017},
  publisher={Public Library of Science San Francisco, CA USA},
  doi={https://doi.org/10.1371/journal.pone.0181742}
}

@article{wu2024pmc,
author = {Wu, Chaoyi and Lin, Weixiong and Zhang, Xiaoman and Zhang, Ya and Xie, Weidi and Wang, Yanfeng},
year = {2024},
month = {04},
pages = {1833-1843},
title = {{PMC-LLaMA}: toward building open-source language models for medicine},
volume = {31},
journal = {Journal of the American Medical Informatics Association},
doi = {10.1093/jamia/ocae045}
}

@inproceedings{kwon-etal-2022-medjex,
    title = "{M}ed{JE}x: A Medical Jargon Extraction Model with {W}iki{'}s Hyperlink Span and Contextualized Masked Language Model Score",
    author = "Kwon, Sunjae  and
      Yao, Zonghai  and
      Jordan, Harmon  and
      Levy, David  and
      Corner, Brian  and
      Yu, Hong",
    editor = "Goldberg, Yoav  and
      Kozareva, Zornitsa  and
      Zhang, Yue",
    booktitle = "Proceedings of the 2022 Conference on Empirical Methods in Natural Language Processing",
    month = dec,
    year = "2022",
    address = "Abu Dhabi, United Arab Emirates",
    publisher = "Association for Computational Linguistics",
    url = "https://aclanthology.org/2022.emnlp-main.805/",
    doi = "10.18653/v1/2022.emnlp-main.805",
    pages = "11733--11751"
}

@inproceedings{huang-etal-2022-understanding,
    title = "Understanding Jargon: Combining Extraction and Generation for Definition Modeling",
    author = "Huang, Jie  and
      Shao, Hanyin  and
      Chang, Kevin Chen-Chuan  and
      Xiong, Jinjun  and
      Hwu, Wen-mei",
    editor = "Goldberg, Yoav  and
      Kozareva, Zornitsa  and
      Zhang, Yue",
    booktitle = "Proceedings of the 2022 Conference on Empirical Methods in Natural Language Processing",
    month = dec,
    year = "2022",
    address = "Abu Dhabi, United Arab Emirates",
    publisher = "Association for Computational Linguistics",
    url = "https://aclanthology.org/2022.emnlp-main.266/",
    doi = "10.18653/v1/2022.emnlp-main.266",
    pages = "3994--4004"
}

@misc{grattafiori2024llama3herdmodels,
      title={The {L}lama 3 Herd of Models}, 
      author={Aaron Grattafiori and Abhimanyu Dubey and Abhinav Jauhri and Abhinav Pandey and Abhishek Kadian and Ahmad Al-Dahle and Aiesha Letman and Akhil Mathur and Alan Schelten and Alex Vaughan and Amy Yang and Angela Fan and Anirudh Goyal and Anthony Hartshorn and Aobo Yang and Archi Mitra and Archie Sravankumar and Artem Korenev and Arthur Hinsvark and Arun Rao and Aston Zhang and Aurelien Rodriguez and Austen Gregerson and Ava Spataru and Baptiste Roziere and Bethany Biron and Binh Tang and Bobbie Chern and Charlotte Caucheteux and Chaya Nayak and Chloe Bi and Chris Marra and Chris McConnell and Christian Keller and Christophe Touret and Chunyang Wu and Corinne Wong and Cristian Canton Ferrer and Cyrus Nikolaidis and Damien Allonsius and Daniel Song and Danielle Pintz and Danny Livshits and Danny Wyatt and David Esiobu and Dhruv Choudhary and Dhruv Mahajan and Diego Garcia-Olano and Diego Perino and Dieuwke Hupkes and Egor Lakomkin and Ehab AlBadawy and Elina Lobanova and Emily Dinan and Eric Michael Smith and Filip Radenovic and Francisco Guzmán and Frank Zhang and Gabriel Synnaeve and Gabrielle Lee and Georgia Lewis Anderson and Govind Thattai and Graeme Nail and Gregoire Mialon and Guan Pang and Guillem Cucurell and Hailey Nguyen and Hannah Korevaar and Hu Xu and Hugo Touvron and Iliyan Zarov and Imanol Arrieta Ibarra and Isabel Kloumann and Ishan Misra and Ivan Evtimov and Jack Zhang and Jade Copet and Jaewon Lee and Jan Geffert and Jana Vranes and Jason Park and Jay Mahadeokar and Jeet Shah and Jelmer van der Linde and Jennifer Billock and Jenny Hong and Jenya Lee and Jeremy Fu and Jianfeng Chi and Jianyu Huang and Jiawen Liu and Jie Wang and Jiecao Yu and Joanna Bitton and Joe Spisak and Jongsoo Park and Joseph Rocca and Joshua Johnstun and Joshua Saxe and Junteng Jia and Kalyan Vasuden Alwala and Karthik Prasad and Kartikeya Upasani and Kate Plawiak and Ke Li and Kenneth Heafield and Kevin Stone and Khalid El-Arini and Krithika Iyer and Kshitiz Malik and Kuenley Chiu and Kunal Bhalla and Kushal Lakhotia and Lauren Rantala-Yeary and Laurens van der Maaten and Lawrence Chen and Liang Tan and Liz Jenkins and Louis Martin and Lovish Madaan and Lubo Malo and Lukas Blecher and Lukas Landzaat and Luke de Oliveira and Madeline Muzzi and Mahesh Pasupuleti and Mannat Singh and Manohar Paluri and Marcin Kardas and Maria Tsimpoukelli and Mathew Oldham and Mathieu Rita and Maya Pavlova and Melanie Kambadur and Mike Lewis and Min Si and Mitesh Kumar Singh and Mona Hassan and Naman Goyal and Narjes Torabi and Nikolay Bashlykov and Nikolay Bogoychev and Niladri Chatterji and Ning Zhang and Olivier Duchenne and Onur Çelebi and Patrick Alrassy and Pengchuan Zhang and Pengwei Li and Petar Vasic and Peter Weng and Prajjwal Bhargava and Pratik Dubal and Praveen Krishnan and Punit Singh Koura and Puxin Xu and Qing He and Qingxiao Dong and Ragavan Srinivasan and Raj Ganapathy and Ramon Calderer and Ricardo Silveira Cabral and Robert Stojnic and Roberta Raileanu and Rohan Maheswari and Rohit Girdhar and Rohit Patel and Romain Sauvestre and Ronnie Polidoro and Roshan Sumbaly and Ross Taylor and Ruan Silva and Rui Hou and Rui Wang and Saghar Hosseini and Sahana Chennabasappa and Sanjay Singh and Sean Bell and Seohyun Sonia Kim and Sergey Edunov and Shaoliang Nie and Sharan Narang and Sharath Raparthy and Sheng Shen and Shengye Wan and Shruti Bhosale and Shun Zhang and Simon Vandenhende and Soumya Batra and Spencer Whitman and Sten Sootla and Stephane Collot and Suchin Gururangan and Sydney Borodinsky and Tamar Herman and Tara Fowler and Tarek Sheasha and Thomas Georgiou and Thomas Scialom and Tobias Speckbacher and Todor Mihaylov and Tong Xiao and Ujjwal Karn and Vedanuj Goswami and Vibhor Gupta and Vignesh Ramanathan and Viktor Kerkez and Vincent Gonguet and Virginie Do and Vish Vogeti and Vítor Albiero and Vladan Petrovic and Weiwei Chu and Wenhan Xiong and Wenyin Fu and Whitney Meers and Xavier Martinet and Xiaodong Wang and Xiaofang Wang and Xiaoqing Ellen Tan and Xide Xia and Xinfeng Xie and Xuchao Jia and Xuewei Wang and Yaelle Goldschlag and Yashesh Gaur and Yasmine Babaei and Yi Wen and Yiwen Song and Yuchen Zhang and Yue Li and Yuning Mao and Zacharie Delpierre Coudert and Zheng Yan and Zhengxing Chen and Zoe Papakipos and Aaditya Singh and Aayushi Srivastava and Abha Jain and Adam Kelsey and Adam Shajnfeld and Adithya Gangidi and Adolfo Victoria and Ahuva Goldstand and Ajay Menon and Ajay Sharma and Alex Boesenberg and Alexei Baevski and Allie Feinstein and Amanda Kallet and Amit Sangani and Amos Teo and Anam Yunus and Andrei Lupu and Andres Alvarado and Andrew Caples and Andrew Gu and Andrew Ho and Andrew Poulton and Andrew Ryan and Ankit Ramchandani and Annie Dong and Annie Franco and Anuj Goyal and Aparajita Saraf and Arkabandhu Chowdhury and Ashley Gabriel and Ashwin Bharambe and Assaf Eisenman and Azadeh Yazdan and Beau James and Ben Maurer and Benjamin Leonhardi and Bernie Huang and Beth Loyd and Beto De Paola and Bhargavi Paranjape and Bing Liu and Bo Wu and Boyu Ni and Braden Hancock and Bram Wasti and Brandon Spence and Brani Stojkovic and Brian Gamido and Britt Montalvo and Carl Parker and Carly Burton and Catalina Mejia and Ce Liu and Changhan Wang and Changkyu Kim and Chao Zhou and Chester Hu and Ching-Hsiang Chu and Chris Cai and Chris Tindal and Christoph Feichtenhofer and Cynthia Gao and Damon Civin and Dana Beaty and Daniel Kreymer and Daniel Li and David Adkins and David Xu and Davide Testuggine and Delia David and Devi Parikh and Diana Liskovich and Didem Foss and Dingkang Wang and Duc Le and Dustin Holland and Edward Dowling and Eissa Jamil and Elaine Montgomery and Eleonora Presani and Emily Hahn and Emily Wood and Eric-Tuan Le and Erik Brinkman and Esteban Arcaute and Evan Dunbar and Evan Smothers and Fei Sun and Felix Kreuk and Feng Tian and Filippos Kokkinos and Firat Ozgenel and Francesco Caggioni and Frank Kanayet and Frank Seide and Gabriela Medina Florez and Gabriella Schwarz and Gada Badeer and Georgia Swee and Gil Halpern and Grant Herman and Grigory Sizov and Guangyi and Zhang and Guna Lakshminarayanan and Hakan Inan and Hamid Shojanazeri and Han Zou and Hannah Wang and Hanwen Zha and Haroun Habeeb and Harrison Rudolph and Helen Suk and Henry Aspegren and Hunter Goldman and Hongyuan Zhan and Ibrahim Damlaj and Igor Molybog and Igor Tufanov and Ilias Leontiadis and Irina-Elena Veliche and Itai Gat and Jake Weissman and James Geboski and James Kohli and Janice Lam and Japhet Asher and Jean-Baptiste Gaya and Jeff Marcus and Jeff Tang and Jennifer Chan and Jenny Zhen and Jeremy Reizenstein and Jeremy Teboul and Jessica Zhong and Jian Jin and Jingyi Yang and Joe Cummings and Jon Carvill and Jon Shepard and Jonathan McPhie and Jonathan Torres and Josh Ginsburg and Junjie Wang and Kai Wu and Kam Hou U and Karan Saxena and Kartikay Khandelwal and Katayoun Zand and Kathy Matosich and Kaushik Veeraraghavan and Kelly Michelena and Keqian Li and Kiran Jagadeesh and Kun Huang and Kunal Chawla and Kyle Huang and Lailin Chen and Lakshya Garg and Lavender A and Leandro Silva and Lee Bell and Lei Zhang and Liangpeng Guo and Licheng Yu and Liron Moshkovich and Luca Wehrstedt and Madian Khabsa and Manav Avalani and Manish Bhatt and Martynas Mankus and Matan Hasson and Matthew Lennie and Matthias Reso and Maxim Groshev and Maxim Naumov and Maya Lathi and Meghan Keneally and Miao Liu and Michael L. Seltzer and Michal Valko and Michelle Restrepo and Mihir Patel and Mik Vyatskov and Mikayel Samvelyan and Mike Clark and Mike Macey and Mike Wang and Miquel Jubert Hermoso and Mo Metanat and Mohammad Rastegari and Munish Bansal and Nandhini Santhanam and Natascha Parks and Natasha White and Navyata Bawa and Nayan Singhal and Nick Egebo and Nicolas Usunier and Nikhil Mehta and Nikolay Pavlovich Laptev and Ning Dong and Norman Cheng and Oleg Chernoguz and Olivia Hart and Omkar Salpekar and Ozlem Kalinli and Parkin Kent and Parth Parekh and Paul Saab and Pavan Balaji and Pedro Rittner and Philip Bontrager and Pierre Roux and Piotr Dollar and Polina Zvyagina and Prashant Ratanchandani and Pritish Yuvraj and Qian Liang and Rachad Alao and Rachel Rodriguez and Rafi Ayub and Raghotham Murthy and Raghu Nayani and Rahul Mitra and Rangaprabhu Parthasarathy and Raymond Li and Rebekkah Hogan and Robin Battey and Rocky Wang and Russ Howes and Ruty Rinott and Sachin Mehta and Sachin Siby and Sai Jayesh Bondu and Samyak Datta and Sara Chugh and Sara Hunt and Sargun Dhillon and Sasha Sidorov and Satadru Pan and Saurabh Mahajan and Saurabh Verma and Seiji Yamamoto and Sharadh Ramaswamy and Shaun Lindsay and Shaun Lindsay and Sheng Feng and Shenghao Lin and Shengxin Cindy Zha and Shishir Patil and Shiva Shankar and Shuqiang Zhang and Shuqiang Zhang and Sinong Wang and Sneha Agarwal and Soji Sajuyigbe and Soumith Chintala and Stephanie Max and Stephen Chen and Steve Kehoe and Steve Satterfield and Sudarshan Govindaprasad and Sumit Gupta and Summer Deng and Sungmin Cho and Sunny Virk and Suraj Subramanian and Sy Choudhury and Sydney Goldman and Tal Remez and Tamar Glaser and Tamara Best and Thilo Koehler and Thomas Robinson and Tianhe Li and Tianjun Zhang and Tim Matthews and Timothy Chou and Tzook Shaked and Varun Vontimitta and Victoria Ajayi and Victoria Montanez and Vijai Mohan and Vinay Satish Kumar and Vishal Mangla and Vlad Ionescu and Vlad Poenaru and Vlad Tiberiu Mihailescu and Vladimir Ivanov and Wei Li and Wenchen Wang and Wenwen Jiang and Wes Bouaziz and Will Constable and Xiaocheng Tang and Xiaojian Wu and Xiaolan Wang and Xilun Wu and Xinbo Gao and Yaniv Kleinman and Yanjun Chen and Ye Hu and Ye Jia and Ye Qi and Yenda Li and Yilin Zhang and Ying Zhang and Yossi Adi and Youngjin Nam and Yu and Wang and Yu Zhao and Yuchen Hao and Yundi Qian and Yunlu Li and Yuzi He and Zach Rait and Zachary DeVito and Zef Rosnbrick and Zhaoduo Wen and Zhenyu Yang and Zhiwei Zhao and Zhiyu Ma},
      year={2024},
      eprint={2407.21783},
      archivePrefix={arXiv},
      primaryClass={cs.AI},
      url={https://arxiv.org/abs/2407.21783}, 
}

@inproceedings{zhang2024ultramedical,
 author = {Zhang, Kaiyan and Zeng, Sihang and Hua, Ermo and Ding, Ning and Chen, Zhang-Ren and Ma, Zhiyuan and Li, Haoxin and Cui, Ganqu and Qi, Biqing and Zhu, Xuekai and Lv, Xingtai and Hu, Jin-Fang and Liu, Zhiyuan and Zhou, Bowen},
 booktitle = {Advances in Neural Information Processing Systems},
 doi = {10.52202/079017-0819},
 editor = {A. Globerson and L. Mackey and D. Belgrave and A. Fan and U. Paquet and J. Tomczak and C. Zhang},
 pages = {26045--26081},
 publisher = {Curran Associates, Inc.},
 title = {UltraMedical: Building Specialized Generalists in Biomedicine},
 url = {https://proceedings.neurips.cc/paper_files/paper/2024/file/2dfc26ce9039f00eee4aba0c54931e46-Paper-Datasets_and_Benchmarks_Track.pdf},
 volume = {37},
 year = {2024}
}

@inproceedings{clark-etal-2019-boolq,
    title = "{B}ool{Q}: Exploring the Surprising Difficulty of Natural Yes/No Questions",
    author = "Clark, Christopher  and
      Lee, Kenton  and
      Chang, Ming-Wei  and
      Kwiatkowski, Tom  and
      Collins, Michael  and
      Toutanova, Kristina",
    editor = "Burstein, Jill  and
      Doran, Christy  and
      Solorio, Thamar",
    booktitle = "Proceedings of the 2019 Conference of the North {A}merican Chapter of the Association for Computational Linguistics: Human Language Technologies, Volume 1 (Long and Short Papers)",
    month = jun,
    year = "2019",
    address = "Minneapolis, Minnesota",
    publisher = "Association for Computational Linguistics",
    url = "https://aclanthology.org/N19-1300/",
    doi = "10.18653/v1/N19-1300",
    pages = "2924--2936"
}

@inproceedings{guo-etal-2024-personalized,
    title = "Personalized Jargon Identification for Enhanced Interdisciplinary Communication",
    author = "Guo, Yue  and
      Chang, Joseph Chee  and
      Antoniak, Maria  and
      Bransom, Erin  and
      Cohen, Trevor  and
      Wang, Lucy  and
      August, Tal",
    editor = "Duh, Kevin  and
      Gomez, Helena  and
      Bethard, Steven",
    booktitle = "Proceedings of the 2024 Conference of the North American Chapter of the Association for Computational Linguistics: Human Language Technologies (Volume 1: Long Papers)",
    month = jun,
    year = "2024",
    address = "Mexico City, Mexico",
    publisher = "Association for Computational Linguistics",
    url = "https://aclanthology.org/2024.naacl-long.255/",
    doi = "10.18653/v1/2024.naacl-long.255",
    pages = "4535--4550"
}

@inproceedings{lucy-etal-2023-words,
    title = "Words as Gatekeepers: Measuring Discipline-specific Terms and Meanings in Scholarly Publications",
    author = "Lucy, Li  and
      Dodge, Jesse  and
      Bamman, David  and
      Keith, Katherine A.",
    editor = "Rogers, Anna  and
      Boyd-Graber, Jordan  and
      Okazaki, Naoaki",
    booktitle = "Findings of the Association for Computational Linguistics: ACL 2023",
    month = jul,
    year = "2023",
    address = "Toronto, Canada",
    publisher = "Association for Computational Linguistics",
    url = "https://aclanthology.org/2023.findings-acl.433/",
    doi = "10.18653/v1/2023.findings-acl.433",
    pages = "6929--6947"
}

@inproceedings{meng2022locating,
 author = {Meng, Kevin and Bau, David and Andonian, Alex and Belinkov, Yonatan},
 booktitle = {Advances in Neural Information Processing Systems},
 doi = {10.52202/068431-1262},
 editor = {S. Koyejo and S. Mohamed and A. Agarwal and D. Belgrave and K. Cho and A. Oh},
 pages = {17359--17372},
 publisher = {Curran Associates, Inc.},
 title = {Locating and Editing Factual Associations in GPT},
 url = {https://proceedings.neurips.cc/paper_files/paper/2022/file/6f1d43d5a82a37e89b0665b33bf3a182-Paper-Conference.pdf},
 volume = {35},
 year = {2022}
}

@misc{marinescu2026medicalinterpretabilityknowledgemaps,
      title={Medical Interpretability and Knowledge Maps of Large Language Models}, 
      author={Razvan Marinescu and Victoria-Elisabeth Gruber and Diego Fajardo},
      year={2026},
      eprint={2510.11390},
      archivePrefix={arXiv},
      primaryClass={cs.LG},
      url={https://arxiv.org/abs/2510.11390}, 
}

@inproceedings{hewitt-liang-2019-designing,
    title = "Designing and Interpreting Probes with Control Tasks",
    author = "Hewitt, John  and
      Liang, Percy",
    editor = "Inui, Kentaro  and
      Jiang, Jing  and
      Ng, Vincent  and
      Wan, Xiaojun",
    booktitle = "Proceedings of the 2019 Conference on Empirical Methods in Natural Language Processing and the 9th International Joint Conference on Natural Language Processing (EMNLP-IJCNLP)",
    month = nov,
    year = "2019",
    address = "Hong Kong, China",
    publisher = "Association for Computational Linguistics",
    url = "https://aclanthology.org/D19-1275/",
    doi = "10.18653/v1/D19-1275",
    pages = "2733--2743"
}

@inproceedings{chang-etal-2024-parts,
    title = "When Parts Are Greater Than Sums: Individual {LLM} Components Can Outperform Full Models",
    author = "Chang, Ting-Yun  and
      Thomason, Jesse  and
      Jia, Robin",
    editor = "Al-Onaizan, Yaser  and
      Bansal, Mohit  and
      Chen, Yun-Nung",
    booktitle = "Proceedings of the 2024 Conference on Empirical Methods in Natural Language Processing",
    month = nov,
    year = "2024",
    address = "Miami, Florida, USA",
    publisher = "Association for Computational Linguistics",
    url = "https://aclanthology.org/2024.emnlp-main.574/",
    doi = "10.18653/v1/2024.emnlp-main.574",
    pages = "10280--10299"
}

@inproceedings{
hendrycks2021measuring,
title={Measuring Massive Multitask Language Understanding},
author={Dan Hendrycks and Collin Burns and Steven Basart and Andy Zou and Mantas Mazeika and Dawn Song and Jacob Steinhardt},
booktitle={International Conference on Learning Representations},
year={2021},
url={https://openreview.net/forum?id=d7KBjmI3GmQ}
}

@misc{zhang2024multiplechoicequestionsefficientrobust,
      title={Multiple-Choice Questions are Efficient and Robust {LLM} Evaluators}, 
      author={Ziyin Zhang and Zhaokun Jiang and Lizhen Xu and Hongkun Hao and Rui Wang},
      year={2024},
      eprint={2405.11966},
      archivePrefix={arXiv},
      primaryClass={cs.CL},
      url={https://arxiv.org/abs/2405.11966}, 
}

@inproceedings{kamalloo-etal-2023-evaluating,
    title = "Evaluating Open-Domain Question Answering in the Era of Large Language Models",
    author = "Kamalloo, Ehsan  and
      Dziri, Nouha  and
      Clarke, Charles  and
      Rafiei, Davood",
    editor = "Rogers, Anna  and
      Boyd-Graber, Jordan  and
      Okazaki, Naoaki",
    booktitle = "Proceedings of the 61st Annual Meeting of the Association for Computational Linguistics (Volume 1: Long Papers)",
    month = jul,
    year = "2023",
    address = "Toronto, Canada",
    publisher = "Association for Computational Linguistics",
    url = "https://aclanthology.org/2023.acl-long.307/",
    doi = "10.18653/v1/2023.acl-long.307",
    pages = "5591--5606"
}

@article{croxford2025evaluating,
  title={Evaluating clinical {AI} summaries with large language models as judges},
  author={Croxford, Emma and Gao, Yanjun and First, Elliot and Pellegrino, Nicholas and Schnier, Miranda and Caskey, John and Oguss, Madeline and Wills, Graham and Chen, Guanhua and Dligach, Dmitriy and Churpek, Matthew and Mayampurath, Anoop and Liao, Frank and Goswami, Cherodeep and Wong, Karen and Patterson, Brian and Afshar, Majid},
  journal={NPJ Digital Medicine},
  volume={8},
  number={1},
  pages={640},
  year={2025},
  publisher={Nature Publishing Group UK London},
  doi = {https://doi.org/10.1038/s41746-025-02005-2}
}

@inproceedings{NEURIPS2019_2c601ad9,
 author = {Michel, Paul and Levy, Omer and Neubig, Graham},
 booktitle = {Advances in Neural Information Processing Systems},
 editor = {H. Wallach and H. Larochelle and A. Beygelzimer and F. d\textquotesingle Alch\'{e}-Buc and E. Fox and R. Garnett},
 pages = {},
 publisher = {Curran Associates, Inc.},
 title = {Are Sixteen Heads Really Better than One?},
 url = {https://proceedings.neurips.cc/paper_files/paper/2019/file/2c601ad9d2ff9bc8b282670cdd54f69f-Paper.pdf},
 volume = {32},
 year = {2019}
}

@inproceedings{10.1145/3488560.3498469,
author = {Ke, Liang and Chen, Xinyu and Wang, Haizhou},
title = {An Unsupervised Detection Framework for Chinese Jargons in the Darknet},
year = {2022},
isbn = {9781450391320},
publisher = {Association for Computing Machinery},
address = {New York, NY, USA},
url = {https://doi.org/10.1145/3488560.3498469},
doi = {10.1145/3488560.3498469},
booktitle = {Proceedings of the Fifteenth ACM International Conference on Web Search and Data Mining},
pages = {458–466},
numpages = {9},
location = {Virtual Event, AZ, USA},
series = {WSDM '22}
}

@inproceedings{10.1145/3473141.3473225,
author = {Lao, Yingying and Zhang, Chenghuan and Wei, Yilun and Han, Dongli},
title = {Detecting and Finding the True Meaning of Jargons},
year = {2021},
isbn = {9781450389723},
publisher = {Association for Computing Machinery},
address = {New York, NY, USA},
url = {https://doi.org/10.1145/3473141.3473225},
doi = {10.1145/3473141.3473225},
booktitle = {Proceedings of the 7th International Conference on Frontiers of Educational Technologies},
pages = {45–50},
numpages = {6},
location = {Bangkok, Thailand},
series = {ICFET '21}
}

@inproceedings{liu-etal-2021-graphine,
    title = "Graphine: A Dataset for Graph-aware Terminology Definition Generation",
    author = "Liu, Zequn  and
      Wang, Shukai  and
      Gu, Yiyang  and
      Zhang, Ruiyi  and
      Zhang, Ming  and
      Wang, Sheng",
    editor = "Moens, Marie-Francine  and
      Huang, Xuanjing  and
      Specia, Lucia  and
      Yih, Scott Wen-tau",
    booktitle = "Proceedings of the 2021 Conference on Empirical Methods in Natural Language Processing",
    month = nov,
    year = "2021",
    address = "Online and Punta Cana, Dominican Republic",
    publisher = "Association for Computational Linguistics",
    url = "https://aclanthology.org/2021.emnlp-main.278/",
    doi = "10.18653/v1/2021.emnlp-main.278",
    pages = "3453--3463"
}

@inproceedings{papandreou-etal-2025-medical,
    title = "Medical Text Simplification From Jargon Detection to Jargon-Aware Prompting",
    author = "Papandreou, Taiki  and
      Bakker, Jan  and
      Kamps, Jaap",
    editor = "Shardlow, Matthew  and
      Alva-Manchego, Fernando  and
      North, Kai  and
      Stodden, Regina  and
      Saggion, Horacio  and
      Khallaf, Nouran  and
      Hayakawa, Akio",
    booktitle = "Proceedings of the Fourth Workshop on Text Simplification, Accessibility and Readability (TSAR 2025)",
    month = nov,
    year = "2025",
    address = "Suzhou, China",
    publisher = "Association for Computational Linguistics",
    url = "https://aclanthology.org/2025.tsar-1.3/",
    doi = "10.18653/v1/2025.tsar-1.3",
    pages = "36--46",
    ISBN = "979-8-89176-176-6"
}

@misc{chen2023meditron70bscalingmedicalpretraining,
      title={{MEDITRON-70B}: Scaling Medical Pretraining for Large Language Models}, 
      author={Zeming Chen and Alejandro Hernández Cano and Angelika Romanou and Antoine Bonnet and Kyle Matoba and Francesco Salvi and Matteo Pagliardini and Simin Fan and Andreas Köpf and Amirkeivan Mohtashami and Alexandre Sallinen and Alireza Sakhaeirad and Vinitra Swamy and Igor Krawczuk and Deniz Bayazit and Axel Marmet and Syrielle Montariol and Mary-Anne Hartley and Martin Jaggi and Antoine Bosselut},
      year={2023},
      eprint={2311.16079},
      archivePrefix={arXiv},
      primaryClass={cs.CL},
      url={https://arxiv.org/abs/2311.16079}, 
}

@inproceedings{wang2023grammar,
 author = {Wang, Bailin and Wang, Zi and Wang, Xuezhi and Cao, Yuan and A. Saurous, Rif and Kim, Yoon},
 booktitle = {Advances in Neural Information Processing Systems},
 doi = {10.52202/075280-2837},
 editor = {A. Oh and T. Naumann and A. Globerson and K. Saenko and M. Hardt and S. Levine},
 pages = {65030--65055},
 publisher = {Curran Associates, Inc.},
 title = {Grammar Prompting for Domain-Specific Language Generation with  Large Language Models},
 url = {https://proceedings.neurips.cc/paper_files/paper/2023/file/cd40d0d65bfebb894ccc9ea822b47fa8-Paper-Conference.pdf},
 volume = {36},
 year = {2023}
}

@inproceedings{yang2023empower,
    title = "Empower Large Language Model to Perform Better on Industrial Domain-Specific Question Answering",
    author = "Yang, Fangkai  and
      Zhao, Pu  and
      Wang, Zezhong  and
      Wang, Lu  and
      Qiao, Bo  and
      Zhang, Jue  and
      Garg, Mohit  and
      Lin, Qingwei  and
      Rajmohan, Saravan  and
      Zhang, Dongmei",
    editor = "Wang, Mingxuan  and
      Zitouni, Imed",
    booktitle = "Proceedings of the 2023 Conference on Empirical Methods in Natural Language Processing: Industry Track",
    month = dec,
    year = "2023",
    address = "Singapore",
    publisher = "Association for Computational Linguistics",
    url = "https://aclanthology.org/2023.emnlp-industry.29/",
    doi = "10.18653/v1/2023.emnlp-industry.29",
    pages = "294--312"
}

@article{10.1145/3764579,
author = {Ling, Chen and Zhao, Xujiang and Lu, Jiaying and Deng, Chengyuan and Zheng, Can and Wang, Junxiang and Chowdhury, Tanmoy and Li, Yun and Cui, Hejie and Zhang, Xuchao and Zhao, Tianjiao and Panalkar, Amit and Mehta, Dhagash and Pasquali, Stefano and Cheng, Wei and Wang, Haoyu and Liu, Yanchi and Chen, Zhengzhang and Chen, Haifeng and White, Chris and Gu, Quanquan and Pei, Jian and Yang, Carl and Zhao, Liang},
title = {Domain Specialization as the Key to Make Large Language Models Disruptive: A Comprehensive Survey},
year = {2025},
issue_date = {February 2026},
publisher = {Association for Computing Machinery},
address = {New York, NY, USA},
volume = {58},
number = {3},
issn = {0360-0300},
url = {https://doi.org/10.1145/3764579},
doi = {10.1145/3764579},
journal = {ACM Comput. Surv.},
month = oct,
articleno = {79},
numpages = {39}
}

@inproceedings{yao-etal-2024-readme,
    title = "{README}: Bridging Medical Jargon and Lay Understanding for Patient Education through Data-Centric {NLP}",
    author = "Yao, Zonghai  and
      Kantu, Nandyala Siddharth  and
      Wei, Guanghao  and
      Tran, Hieu  and
      Duan, Zhangqi  and
      Kwon, Sunjae  and
      Yang, Zhichao  and
      Yu, Hong",
    editor = "Al-Onaizan, Yaser  and
      Bansal, Mohit  and
      Chen, Yun-Nung",
    booktitle = "Findings of the Association for Computational Linguistics: EMNLP 2024",
    month = nov,
    year = "2024",
    address = "Miami, Florida, USA",
    publisher = "Association for Computational Linguistics",
    url = "https://aclanthology.org/2024.findings-emnlp.737/",
    doi = "10.18653/v1/2024.findings-emnlp.737",
    pages = "12609--12629"
}

@inproceedings{song-etal-2023-matsci,
    title = "{M}at{S}ci-{NLP}: Evaluating Scientific Language Models on Materials Science Language Tasks Using Text-to-Schema Modeling",
    author = "Song, Yu  and
      Miret, Santiago  and
      Liu, Bang",
    editor = "Rogers, Anna  and
      Boyd-Graber, Jordan  and
      Okazaki, Naoaki",
    booktitle = "Proceedings of the 61st Annual Meeting of the Association for Computational Linguistics (Volume 1: Long Papers)",
    month = jul,
    year = "2023",
    address = "Toronto, Canada",
    publisher = "Association for Computational Linguistics",
    url = "https://aclanthology.org/2023.acl-long.201/",
    doi = "10.18653/v1/2023.acl-long.201",
    pages = "3621--3639"
}

@article{10.1093/nar/gkh061,
    author = {Bodenreider, Olivier},
    title = {The {U}nified {M}edical {L}anguage {S}ystem ({UMLS}): integrating biomedical terminology},
    journal = {Nucleic Acids Research},
    volume = {32 Database issue},
    pages = {D267--D270},
    year = {2004},
    month = {01},
    issn = {0305-1048},
    doi = {10.1093/nar/gkh061},
    url = {https://doi.org/10.1093/nar/gkh061},
    eprint = {https://academic.oup.com/nar/article-pdf/32/suppl_1/D267/7621558/gkh061.pdf},
}

@inproceedings{reimers-2019-sentence-bert,
    title = "Sentence-{BERT}: Sentence Embeddings using {S}iamese {BERT}-Networks",
    author = "Reimers, Nils  and
      Gurevych, Iryna",
    editor = "Inui, Kentaro  and
      Jiang, Jing  and
      Ng, Vincent  and
      Wan, Xiaojun",
    booktitle = "Proceedings of the 2019 Conference on Empirical Methods in Natural Language Processing and the 9th International Joint Conference on Natural Language Processing (EMNLP-IJCNLP)",
    month = nov,
    year = "2019",
    address = "Hong Kong, China",
    publisher = "Association for Computational Linguistics",
    url = "https://aclanthology.org/D19-1410/",
    doi = "10.18653/v1/D19-1410",
    pages = "3982--3992"
}

@inproceedings{ylonen22,
    title = "Wiktextract: {W}iktionary as Machine-Readable Structured Data",
    author = "Ylonen, Tatu",
    editor = "Calzolari, Nicoletta  and
      B{\'e}chet, Fr{\'e}d{\'e}ric  and
      Blache, Philippe  and
      Choukri, Khalid  and
      Cieri, Christopher  and
      Declerck, Thierry  and
      Goggi, Sara  and
      Isahara, Hitoshi  and
      Maegaard, Bente  and
      Mariani, Joseph  and
      Mazo, H{\'e}l{\`e}ne  and
      Odijk, Jan  and
      Piperidis, Stelios",
    booktitle = "Proceedings of the Thirteenth Language Resources and Evaluation Conference",
    month = jun,
    year = "2022",
    address = "Marseille, France",
    publisher = "European Language Resources Association",
    url = "https://aclanthology.org/2022.lrec-1.140/",
    pages = "1317--1325"
}

@misc{guo2017calibrationmodernneuralnetworks,
      title={On Calibration of Modern Neural Networks}, 
      author={Chuan Guo and Geoff Pleiss and Yu Sun and Kilian Q. Weinberger},
      year={2017},
      eprint={1706.04599},
      archivePrefix={arXiv},
      primaryClass={cs.LG},
      url={https://arxiv.org/abs/1706.04599}, 
}

@article{jiang-etal-2021-know,
    title = "How Can We Know When Language Models Know? On the Calibration of Language Models for Question Answering",
    author = "Jiang, Zhengbao  and
      Araki, Jun  and
      Ding, Haibo  and
      Neubig, Graham",
    editor = "Roark, Brian  and
      Nenkova, Ani",
    journal = "Transactions of the Association for Computational Linguistics",
    volume = "9",
    year = "2021",
    address = "Cambridge, MA",
    publisher = "MIT Press",
    url = "https://aclanthology.org/2021.tacl-1.57/",
    doi = "10.1162/tacl_a_00407",
    pages = "962--977"
}

@inproceedings{vigna2015weighted,
author = {Vigna, Sebastiano},
title = {A Weighted Correlation Index for Rankings with Ties},
year = {2015},
isbn = {9781450334693},
publisher = {International World Wide Web Conferences Steering Committee},
address = {Republic and Canton of Geneva, CHE},
url = {https://doi.org/10.1145/2736277.2741088},
doi = {10.1145/2736277.2741088},
booktitle = {Proceedings of the 24th International Conference on World Wide Web},
pages = {1166–1176},
numpages = {11},
location = {Florence, Italy},
series = {WWW '15}
}

@article{naeini15, title={Obtaining Well Calibrated Probabilities Using Bayesian Binning}, volume={29}, url={https://ojs.aaai.org/index.php/AAAI/article/view/9602}, DOI={10.1609/aaai.v29i1.9602}, abstractNote={ &amp;lt;p&amp;gt; Learning probabilistic predictive models that are well calibrated is critical for many prediction and decision-making tasks in artificial intelligence. In this paper we present a new non-parametric calibration method called Bayesian Binning into Quantiles (BBQ) which addresses key limitations of existing calibration methods. The method post processes the output of a binary classification algorithm; thus, it can be readily combined with many existing classification algorithms. The method is computationally tractable, and empirically accurate, as evidenced by the set of experiments reported here on both real and simulated datasets. &amp;lt;/p&amp;gt; }, number={1}, journal={Proceedings of the AAAI Conference on Artificial Intelligence}, author={Naeini, Mahdi Pakdaman and Cooper, Gregory and Hauskrecht, Milos}, year={2015}, month={Feb.} }

\newpage

\appendix

% \section{Appendix}
% \label{sec:appendix}

\section{Computational setup}
 All SBERT-based components of our pipeline — the Wiktionary-based jargon filtering (Appendix~\ref{app:wiktionary-tags}), the within-term semantic deduplication (Appendix~\ref{app:readme-detail}), and the distractor ranking for the JU benchmark (Section~\ref{sec:ju-task}) — use the all-mpnet-base-v2 checkpoint from the sentence-transformers library. Embeddings are compared using cosine similarity. SBERT encoding runs on a single NVIDIA RTX A6000 GPU and takes under 5 minutes for the full README-exp-good filtering and JU distractor ranking.

All experiments use Llama-3.1-8B-Instruct and Llama-3.1-8B-UltraMedical. Both models are used in inference-only settings: The decomposition procedure caches per-component projections to disk, and the reweighting procedure optimizes 1056 scalar weights against these cached projections on CPU, without further forward passes through the 8B models. Decomposition runs on two NVIDIA RTX A6000 GPUs (48GB each) and take approximately 5 minutes for JU and 8 minutes for JI per model. Cumulative ablation experiments (top-k, random, and cross-model variants across both tasks and both models) total approximately 2 hours of GPU time. Reweighting optimization on cached projections takes under one minute per task on CPU.

\section{Decomposition setup}
Our JU and JI tasks are zero-shot evaluations with a single canonical prompt template per task. 
We use the publicly released code from \citet{chang-etal-2024-parts}, with the following arguments held constant across all experiments:

\begin{itemize}
\item \texttt{n\_shots}: 0 (zero-shot evaluation)
\item \texttt{format}: 1 (single canonical prompt template per task)
\item \texttt{seed\_list}: [0] (single seed; see discussion below)
\end{itemize}

In our zero-shot setting, the \texttt{seed\_list} argument passed to the decomposition and evaluation scripts does not affect results, the forward pass is deterministic and no demonstration sampling occurs at evaluation time. We use the default value for all other hyperparameters.

\begin{table*}[t]
\small
    \centering
    \renewcommand{\arraystretch}{1.3}
    % The middle column is now right-aligned
    \begin{tabularx}{\textwidth}{@{} l >{\centering\arraybackslash}X r @{}}
        \textbf{Dataset} & \textbf{Data Description} & \textbf{Size} \\
        \midrule
        \multicolumn{3}{l}{\textit{Medical Domain}} \\
        \midrule

        README-exp-good & Unfiltered medical jargon data & 113,659 \\
        Jargon Dataset $\mathcal{J}$ & SBERT filtered jargon & 5,977 \\
        Non-Jargon Dataset $\mathcal{N}$ & SBERT filtered non-jargon & 2,095 \\
        Train data & 70\% split from $\mathcal{J}$ & 4,183 \\
        Test data & 25\% split from $\mathcal{J}$ & 1,495 \\
        Dev data & 5\% split from $\mathcal{J}$ & 299 \\
        \midrule
        \multicolumn{3}{l}{\textit{Material Science Domain}} \\
        \midrule
        Mat-JI & Material Science Jargon Identification task & 2,990 \\
        \midrule
        \multicolumn{3}{l}{\textit{Others}} \\
        \midrule
        BoolQ & Control group to isolate out non-jargon components & 2,990 \\

    \end{tabularx}
    \caption{ Data statistics for the medical-domain jargon benchmark. Rows show the unfiltered README-exp-good source, the SBERT-filtered jargon ($\mathcal{J}$) and non-jargon ($\mathcal{N}$) sets, and the Train/Test/Dev splits derived from $\mathcal{J}$.}
    \label{tab:model_stats}
\end{table*}   

\section{Details of README}
\label{app:readme-detail}
README~\cite{yao-etal-2024-readme} provides multiple subsets that differ in how their general definitions are sourced. The README-exp variants contain general definitions retrieved from UMLS, an authoritative biomedical vocabulary maintained by the National Library of Medicine. The README-syn variants, in contrast, contain general definitions generated by GPT-3.5-turbo through README's data augmentation pipeline. Within the expert-annotated branch, $\texttt{README-exp\_good}$ is the subset whose UMLS-retrieved general definitions passed README's Examiner stage, in which an LLM-based quality check filters out definitions deemed inaccurate, contextually misaligned, or otherwise unsuitable for the corresponding lay definition.

We only use $\texttt{README-exp\_good}$ as our jargon source. Restricting our benchmark to UMLS-retrieved definitions ensures that every jargon term in our evaluation is associated with a definition from an authoritative medical knowledge base rather than from a language model.

\section{Wiktionary-based definition filtering}
\label{app:wiktionary-tags}
Section~\ref{sec:section-3README} describes how Wiktionary tags are used to identify the conventional (non-medical) senses of each candidate jargon term. 
Let $\mathcal{T}_\text{tech}$ denote the set of Wiktionary tags that indicate specialized technical usage. For each term $w$ in \texttt{README-exp\_good}, we retrieve its set of Wiktionary definitions $D_{wik}$. We define the subset of conventional (non-medical) definitions as:
    \begin{equation}
    D_{conv} = \{d \in D_{wik} \mid \text{tags}(d) \cap \mathcal{T}_\text{tech} = \emptyset\}\;
    \end{equation}
     We retain $w$ as jargon if its README medical definition $d_w$ is sufficiently distant from all conventional senses:
\begin{multline}
    \mathcal{J} = \{ w \in \mathcal{T} : \phi(w) < \tau \}, \\
 \text{where } \phi(w) = \max_{d \in D_{\text{conv}}} \mathrm{sim}(f_\theta(d_w), f_\theta(d))
\end{multline}

    \noindent Where $f_{\theta}$ represents the SBERT embedding, $\text{sim}(\cdot)$ is the cosine similarity function, and $\tau = 0.5$ chosen by manual inspection. Terms whose medical definition differs sufficiently from their non-medical senses (similarity $< \tau$) are retained as jargon. Terms that fail this criterion are reassigned to the non-jargon set. 

The tag list $\mathcal{T}_\text{tech}$ is constructed by inspecting the Wiktionary tags that appeared most frequently across the candidate terms in \texttt{README-exp\_good} and manually selecting those that indicate medical or technical usage. 

The set $\mathcal{T}_\text{tech}$ comprises the following tags: jargon, slang, medicine, pathology, pharmacology, anatomy, surgery, immunology, physiology, biology, microbiology, biochemistry, genetics, zoology, botany, entomology, psychology.

\section{Data Statistics}
\label{app:data-stats}
Table \ref{tab:model_stats} shows the data description and the data size of each of the benchmarks. The unfiltered \texttt{README-exp-good} source contains 113,659 entries. 

We first normalize both the jargon term and its general definition using regular expressions. We then drop duplicate (jargon, general definition) pairs on the normalized strings, retaining a single representative entry per unique pair. Through manual inspection, we identified many jargon entries with multiple definitions, but these definitions often differ only in word order. To remove these, we apply a within-term semantic de-duplication step using SBERT. For each jargon term, we encode all of its remaining general definitions and compute pairwise cosine similarities. We then perform a greedy selection: iterating through definitions in their original order, we retain a candidate only if its cosine similarity to every already-retained definition for that term falls below a threshold of $\tau$ = 0.92. The threshold is chosen manually, which results in 8,072 remaining entries. Finally, we use SBERT to perform the Wiktionary-based filtering described in Appendix~\ref{app:wiktionary-tags}.

\section{Example prompt for JU}
\label{app:ju-example}
Table~\ref{tab:med-ju-example} shows an example prompt of JU.
\begin{table}[t]
    \centering
    \renewcommand{\arraystretch}{1.5}
    \small

    \begin{tabular}{|p{0.95\linewidth}|}
        \hline
        \textbf{System Prompt:} \\
        You are a clinical assistant. Given a medical TERM (jargon), an optional EHR context, and several candidate general definitions, select the single best general definition. Reply with ONE letter only (A, B, C, D, or E) \\
        \hline
        \textbf{User Input:} \\
        TERM: aorta \\
        EHR CONTEXT: Over the guidewire, a 5-French Sos catheter was reshaped in the descending thoracic aorta and utilized to catheterize an iliolumbar branch extending from the infrarenal abdominal aorta. \\
        \\
        Choose the best general definition: \\
        A) The main trunk of the systemic arteries. \\
        B) The continuation of the subclavian artery; it distributes over the upper limb, axilla, chest and shoulder. \\
        C) The venous trunk which returns blood from the head, neck, upper extremities and chest. \\
        D) The venous trunk which receives blood from the lower extremities and from the pelvic and abdominal organs. \\
        E) An arterial trunk that contains parts of the posterior tibial artery and fibular artery. \\
        \\
        Answer with a single letter only. \\
        \hline
    \end{tabular}
    \caption{Example prompt of JU}
    \label{tab:med-ju-example}
\end{table}

\section{Example prompt for JI}
\label{app:ji-example}
Table~\ref{tab:med-ji-example} shows an example prompt of Medical JI.
\begin{table}[t]
    \centering
    \renewcommand{\arraystretch}{1.5}
    \small
    \begin{tabular}{|p{0.95\linewidth}|}
        \hline
        \textbf{System Prompt:} \\
        You are a clinical assistant. Given a TERM from an EHR (Electronic Health Record) note and its context, determine whether the term is medical jargon — specialized medical terminology that a layperson would likely not understand. Answer only with a single letter (A or B). \\
        \hline
        \textbf{User Input:} \\
        TERM: stable \\
        EHR CONTEXT: The rest of his issues, including his hepatic encephalopathy, hypertension, and hypernutrition, are stable, and he appears to be eating adequately on his own, so we do not need to reinsert his oral gastric tube again. \\
        \\
        Is this term a medical jargon (specialized terminology a layperson would not understand)? \\
        A) Yes \\
        B) No \\
        Answer with A or B only. \\
        \hline
    
    \end{tabular}
    \caption{Example prompt of Medical JI}
    \label{tab:med-ji-example}
\end{table}

\section{Prompt Sensitivity Analysis}
To evaluate whether the performance gap between Llama-3.1-8B-Instruct and Llama-3.1-8B-UltraMedical is sensitive to prompt formulation, we evaluate both models under several variants of the canonical prompts used for Jargon Understanding (JU) and Jargon Identification (JI). Each variant modifies a single aspect of the original prompt while keeping the benchmark examples, model settings, and evaluation procedure fixed. For JI, we vary the system persona, question wording, definition of jargon, and answer-label ordering. For JU, we vary the system persona and question wording. We do not separately shuffle the JU answer options because the correct answer is already distributed uniformly across labels A--E by construction. The canonical prompts in Appendices~\ref{app:ju-example} and \ref{app:ji-example} serve as reference points for the variants described below.

\subsection{Results}
Table~\ref{tab:ji_prompt_sensitivity} reports JI accuracy across prompt variants. The relative ordering of the two models is consistent across all tested prompts: Llama-3.1-8B-Instruct outperforms UltraMedical under each variant, although absolute accuracy varies with prompt wording.
\begin{table}[t]
\small
\centering
\resizebox{\columnwidth}{!}{%
\begin{tabular}{lrrrrr}
\toprule
\textbf{Model} & \textbf{Original} & \textbf{Persona} &
\textbf{Question R.} & \textbf{Definition R.} & \textbf{Ordering} \\
\midrule
Instruct      & 58.8 & 58.0 & 60.3 & 60.2 & 58.1 \\
UltraMedical & 53.1 & 53.2 & 51.0 & 51.5 & 53.9 \\
\bottomrule
\end{tabular}%
}
\caption{JI prediction accuracy across prompt variants. “R.” denotes rewording (e.g., Question R. = question rewording; Definition R. = definition rewording).}
\label{tab:ji_prompt_sensitivity}
\end{table}

Table~\ref{tab:ju_prompt_sensitivity} reports the corresponding analysis for JU. Instruct likewise maintains higher accuracy than UltraMedical across all tested variants.

\begin{table}[t]
\small
\centering
\begin{tabular}{l r r r r r} 
\textbf{Model} & \textbf{Original} & \textbf{Persona}& \textbf{Question R.} \\ 
\midrule
Instruct & 76.1 &  74.7 & 76.3\\
UltraMedical & 74.1 & 74.4 & 74.5\\
\end{tabular}
\caption{JU prediction accuracy across prompt variants.}
\label{tab:ju_prompt_sensitivity}
\end{table}

These results indicate that the observed performance ordering between the two models is qualitatively robust to the prompt variations considered. We do not claim invariance to arbitrary prompting strategies; Rather, the tested variations suggest that the main comparison is not specific to the canonical wording used in our main experiments.

\subsection{Prompt Variants.}
This section shows examples of prompt variants tested.
\paragraph{Persona Removal.} This variant removes the clinical-assistant persona while leaving the remaining prompt unchanged. This is performed on both JI and JU prompts. Table~\ref{tab:JI-ex-persona} shows an example prompt.

\begin{table}[t]
    \centering
    \renewcommand{\arraystretch}{1.5}
    \small
    \begin{tabular}{|p{0.95\linewidth}|}
        \hline
        \textbf{Original System Prompt:} \\
        You are a clinical assistant. Given a TERM from an EHR
(Electronic Health Record) note and its context, determine
whether the term is medical jargon — specialized medical
terminology that a layperson would likely not understand.
Answer only with a single letter (A or B). \\
        \hline
        \textbf{Modified System Input:} \\
        \textcolor{red}{(Removed)} Given a TERM from an EHR
(Electronic Health Record) note and its context, determine
whether the term is medical jargon — specialized medical
terminology that a layperson would likely not understand.
Answer only with a single letter (A or B). \\
        \hline
    \end{tabular}
    \caption{Example prompt of JI with persona removal.}
    \label{tab:JI-ex-persona}
\end{table}

\paragraph{Question rewording.} The final question is replaced, while all other prompt components remain unchanged. This is performed on both JI and JU. Table~\ref{tab:JI-ex-question} shows an example JI prompt, and Table~\ref{tab:JU-ex-question} shows an example JU prompt.

\begin{table}[t]
    \centering
    \renewcommand{\arraystretch}{1.5}
    \small
    \begin{tabular}{|p{0.95\linewidth}|}
        \hline
        \textbf{Original User Prompt:} \\
         Is this term a medical jargon (specialized terminology a layperson would not understand)? \\
        \hline
        \textbf{Modified User Prompt:} \\
        \textcolor{red}{Would a non-expert reader most likely not understand this term here?} \\
        \hline
    \end{tabular}
    \caption{Example prompt of JI with question rewording.}
    \label{tab:JI-ex-question}
\end{table}

\begin{table}[!htbp]
    \centering
    \renewcommand{\arraystretch}{1.5}
    \small
    \begin{tabular}{|p{0.95\linewidth}|}
        \hline
        \textbf{Original User Prompt:} \\
        TERM: aorta \\
        EHR CONTEXT: Over the guidewire, a 5-French Sos catheter was reshaped in the descending thoracic aorta and utilized to catheterize an iliolumbar branch extending from the infrarenal abdominal aorta. \\
        \\
        Choose the best general definition:\\
        \hline
        \textbf{Modified User Input:} \\
        TERM: aorta \\
        EHR CONTEXT: Over the guidewire, a 5-French Sos catheter was reshaped in the descending thoracic aorta and utilized to catheterize an iliolumbar branch extending from the infrarenal abdominal aorta. \\
        \\
        \textcolor{red}{Choose the definition that best matches the term as used in the context: } \\
        \hline
    \end{tabular}
    \caption{Example prompt of JU with question rewording.}
    \label{tab:JU-ex-question}
\end{table}

\paragraph{Definition rewording.} The definition of jargon is changed from \textit{specialized medical terminology that a layperson would likely not understand} to \textit{a term that a person without medical training would likely not understand.}
Table~\ref{tab:JI-ex-instruction} shows an example prompt.

\begin{table}[t]
    \centering
    \renewcommand{\arraystretch}{1.5}
    \small
    \begin{tabular}{|p{0.95\linewidth}|}
        \hline
        \textbf{Original System Prompt:} \\
          You are a clinical assistant.Given a TERM from an EHR
(Electronic Health Record) note and its context, determine
whether the term is medical jargon — specialized medical
terminology that a layperson would likely not understand.
Answer only with a single letter (A or B).\\
        \hline
        \textbf{Modified System Prompt:} \\
         You are a clinical assistant.Given a TERM from an EHR (Electronic Health Record) note and its context, determine whether the term is medical jargon — \textcolor{red}{a term that a person without medical training would likely not understand.} Answer only with a single letter (A or B).\\
        \hline
    \end{tabular}
    \caption{Example prompt of JI with definition rewording.}
    \label{tab:JI-ex-instruction}
\end{table}

\paragraph{Answer label ordering.} The semantic content of the prompt is unchanged, but the order of the binary response options is reversed. Table~\ref{tab:JI-ex-reorder} shows an example prompt.

\begin{table}[t]
    \centering
    \renewcommand{\arraystretch}{1.5}
    \small
    \begin{tabular}{|p{0.95\linewidth}|}
        \hline
        \textbf{Original User Prompt:} \\
         Is this term a medical jargon (specialized terminology a layperson would not understand)? \\
        A) Yes \\
        B) No \\
        \hline
        \textbf{Modified User Prompt:} \\
        Is this term a medical jargon (specialized terminology a layperson would not understand)? \\
        \textcolor{red}{A) No} \\
        \textcolor{red}{B) Yes} \\
        
        \hline
    \end{tabular}
    \caption{Example prompt of JI with label reordering.}
    \label{tab:JI-ex-reorder}
\end{table}

\section{Construction of Mat-JI}
\label{app:mat-ji-app}
This section provides the full construction pipeline for Mat-JI, the materials science jargon identification benchmark used in our cross-domain analysis. MatScholar is a named entity recognition (NER) corpus consisting of materials science paper abstracts annotated with BIO tags for domain entities such as materials, properties, applications, and synthesis methods. Each token is labeled B- (beginning of an entity span), I- (inside an entity span), or O (outside any entity). Because MatScholar provides only NER annotations and no dictionary definitions, the SBERT-based filtering procedure used for the medical pipeline cannot be applied here. We therefore construct the benchmark directly from the BIO structure.

\paragraph{Positive Class.} Each contiguous span of B-/I- tagged tokens is treated as a single jargon term, since these spans correspond to recognized domain entities by construction. 

\paragraph{Negative Class.} The naïve approach would be to sample negatives uniformly from O-tagged tokens. However, this yields trivial distractors: stop words (e.g., the, and), basic academic verbs (e.g., study, show), and other surface-level non-entities that any model can trivially classify. Such negatives would inflate accuracy without testing the model's ability to distinguish specialized terminology from general scientific vocabulary. To construct a more challenging negative class, we use an LLM-as-a-judge procedure to mine O-tagged tokens that are not materials science entities but are perceived as technical.

\paragraph{LLM annotation framework}
We use Claude Sonnet 4.5 as the judge model. The judge is prompted to act as an expert computational linguist and rate each O-tagged token on a 1–5 ordinal scale of perceived technical complexity. We prompt Claude Sonnet 4.5 using the prompt shown in Table~\ref{tab:prompt-matji}.

\begin{table}[t]
    \begin{tabular}{|p{0.95\linewidth}|}

        \hline
        Based on the following term from a materials science abstract, how would you rate its perceived technical complexity?
        The term should be evaluated as a non-material science term. We are looking for terms that may sound technical but are not themselves materials science entities.

        \vspace{4pt}
        \textbf{Context:} [Abstract]

        \textbf{Term:} [O-tokens]

        \vspace{4pt}
        On a scale of 1 to 5, where:\\
        1 = Basic English (e.g., \emph{the}, \emph{however}, \emph{because})\\
        2 = Standard academic vocabulary (e.g., \emph{study}, \emph{analyze}, \emph{method})\\
        3 = Broad scientific vocabulary (e.g., \emph{temperature}, \emph{velocity}, \emph{concentration})\\
        4 = College-level technical vocabulary (e.g., \emph{synthesized}, \emph{configuration}, \emph{empirical})\\
        5 = Deceptive hard negative: mimics specialized nomenclature or is jargon from another field, but is not a materials science entity (e.g., \emph{stochastic}, \emph{orthogonal}, \emph{eigenvector})

        \vspace{4pt}
        Please only respond with a number from 1 to 5. \\
        \hline
    \end{tabular}
    \caption{Prompt used for negative word annotation in Mat-JI.}
    \label{tab:prompt-matji}
\end{table}

We sample negatives uniformly across scores 3, 4, and 5 rather than from score 5 alone. This shifts the bulk of the benchmark away from boundary cases while still ensuring meaningful difficulty.
%provides a realistic gradient of scientific vocabulary against which to evaluate jargon recognition. 
The stratified negatives are combined with the positive entity spans at a 1:1 ratio to produce a balanced 2,990-item Mat-JI evaluation set, matching the size of the medical JI test set. Table~\ref{tab:mat-JI-ex} shows an example of Mat-JI.

\begin{table}[t]
    \centering
    \renewcommand{\arraystretch}{1.5}
    \small
    \begin{tabular}{|p{0.95\linewidth}|}
        \hline
        \textbf{System Prompt:} \\
        You are a research assistant. Given a TERM from a material science paper abstract and its context, determine whether the term is material science jargon — specialized terminology that a reader outside the field would likely not understand. Answer only with a single letter (A or B). \\
        \hline
        \textbf{User Input:} \\
        TERM: couples \\
        ABSTRACT CONTEXT: analysis of the temperature dependence of the elastic modulus indicates that in the particular range of carrier concentration there exists a narrow electronic band in the vicinity of the fermi level which couples to the shearing strain exx - eyy . \\
        \\
        Is this term a material science jargon (specialized terminology a non-expert would not understand)? \\
        A) Yes \\
        B) No \\
        Answer with A or B only. \\
        \hline
        \textbf{Assistant Output (Gold Label):} \\
        B \\
        \hline
    \end{tabular}
    \caption{Example prompt of Mat-JI.}
    \label{tab:mat-JI-ex}
\end{table}

\section{Examples of UltraMedical's errors}
\label{app:ultramed-error-example}
Table~\ref{tab:med-ji-overclassification} shows three error examples where UltraMedical misclassified a non-jargon term as jargon, while Instruct correctly classified it as non-jargon. Manual inspection of these cases reveals two recurring categories.

The first category consists of everyday English words that appear in clinical documentation but are not themselves specialized medical terminology. Words such as \textit{elevated}, and \textit{consumption} are widely understood outside clinical settings, but UltraMedical systematically treats their appearance in an EHR context as evidence of jargon status. The second category consists of lay-accessible medical terms, that is, words that are technically medical in origin but have become part of everyday health literacy. Examples include \textit{vaccination} and \textit{hospitalizations}, which the model overestimates as specialized terminology despite being broadly familiar to non-experts.

Together, these patterns suggest that UltraMedical's decision boundary is anchored more to the clinical context in which a term appears than to the term's intrinsic semantics. Fine-tuning on medical text, where the majority of training examples involve genuine specialized terminology embedded in clinical contexts, plausibly strengthens this contextual association at the expense of term-level discrimination.

\begin{table}[t]
    \centering

    \small % Slightly smaller text makes it look much more professional
    \setlength{\tabcolsep}{4pt} % Tightens the horizontal padding
    
    % Use 'X' for the question to fill space
    % Use 'l' (left) or 'c' (center) for the numbers so they don't wrap strangely
    \begin{tabularx}{\columnwidth}{>{\raggedright\arraybackslash}X r r}
        \textbf{Question} & \textbf{Logit (Jargon)} & \textbf{Logit (Non-Jargon)} \\ 
        \midrule
        
        This is likely a small region of ischemia caused by \textbf{elevated} demand in the context of the current illness, which should resolve without incident \dots & 29.38 & 23.5 \\ 
        \midrule
        
        DISCHARGE DIAGNOSIS: Chronic obstructive asthma with exacerbation, obstructive bronchitis, a history of noncompliance with medical treatment, and the necessity for prophylactic \textbf{vaccination} and inoculation against respiratory viruses. & 29.75 & 24.0 \\ 
        \midrule
        
        Longitudinal incisions were \textbf{simultaneously} made over the PIP joints of 2, 3, 4, and 5, and capsulotomies of the PIP joints were performed. & 30.13 & 20.13 \\ 
    \end{tabularx}
    \caption{Logit scores for the jargon and non-jargon labels on two examples where the Instruct model correctly classified the bolded term as non-jargon but UltraMedical misclassified it as jargon. The logit scores shown are from the UltraMedical model.}
    \label{tab:med-ji-overclassification}
\end{table}

\section{Examples of Instruct's errors}
\label{app:instruct-error-example}
\begin{table}[t]
    \centering
    \small % Slightly smaller text makes it look much more professional
    \setlength{\tabcolsep}{4pt} % Tightens the horizontal padding
    \renewcommand{\arraystretch}{1.8} % Increases space between rows
    
    % Use 'X' for the question to fill space
    % Use 'l' (left) or 'c' (center) for the numbers so they don't wrap strangely
    \begin{tabularx}{\columnwidth}{>{\raggedright\arraybackslash}X r r}
        \textbf{Question} & \textbf{Logit (Jargon)} & \textbf{Logit (Non-Jargon)} \\ 
        \midrule
        
         Overlying surgical drain, skin staples, and \textbf{dressings} are applied. \dots & 33.5 & 37.5 \\ 
        \midrule
        
         Due to the patient's element of chronic venostasis disease, approximately 1 week of \textbf{compression} therapy was initiated prior to the surgery to reduce the edema. & 33.75 & 34.25 \\  
    \end{tabularx}
    \caption{Logit scores for the jargon and non-jargon labels on two examples where the Instruct model incorrectly classified the bolded term as non-jargon but UltraMedical classified it as jargon. The logit scores shown are from the Instruct model.}
    \label{tab:gen-ji-examples}
\end{table}
For completeness, we also conducted a qualitative analysis of the 196 disagreement cases where the Instruct model misclassified a jargon term as non-jargon while UltraMedical correctly identified it as jargon. Instruct's errors reflect the opposite tendency: The model interprets terms according to their conventional meanings and is insufficiently sensitive to the specialized senses they acquire in clinical contexts.

Table~\ref{tab:gen-ji-examples} illustrates this pattern with two representative examples. In the first, \textit{dressings} appears in a post-surgical context where it refers to wound coverings, but the Instruct model assigns a higher logit to the non-jargon label, defaulting to the term's everyday culinary sense. The second example shows the same pattern for \textit{compression}, which denotes a specific therapeutic protocol for venous disease in this clinical setting rather than the general physical act of squeezing. In both cases, the logit margins are narrow, suggesting that the model recognizes some clinical signal but ultimately weights the conventional sense more heavily.

Instruct maintains a more conservative decision boundary, requiring stronger contextual signals before classifying a term as jargon. This conservatism results in a reversed error pattern compared to UltraMedical: Rather than over-extending the jargon label to terms appearing in clinical contexts, Instruct fails to flag genuinely specialized usages when the surface form of the term remains familiar.

\section{Top performing components on both tasks for both models}
\label{app:top-performing-comp}
Table~\ref{tab:head_rank_JU} and~\ref{tab:head_rank_JI}  show the top 5 performing components of the two models on JU and JI, respectively.

On JU, the same four components — L17H24, L26H3, MLP-29, and L31H1 — appear in the top-5 of both models, differing only in rank order, with individual accuracies approaching but not exceeding full-model performance (e.g., 75.5\% vs.\ 76.1\% for Instruct). On JI, by contrast, the top-5 lists share no components between the two models. However, the individual top components exceed full-model accuracy in both cases (61.3\% vs.\ 58.8\% for Instruct; 58.6\% vs.\ 53.1\% for UltraMedical). This pattern is consistent with~\citet{chang-etal-2024-parts}'s observation that individual components can outperform full models on tasks where the aggregate output reflects competing signals, and reinforces the finding from Section~\ref{sec:comp-analysis} that JI relies on distributed contributions whose ranking is sensitive to fine-tuning, whereas JU is supported by a small set of specialized components shared across models.

\begin{table}[t]
\centering
\small
\setlength{\tabcolsep}{4pt}
\begin{tabular}{l r r r r} 
\textbf{Rank} & \textbf{Instruct} & \textbf{Accuracy} & \textbf{UltraMedical} & \textbf{Accuracy} \\ 
\midrule
1 & L26H3 & 75.52\% & L31H1 & 72.37\% \\ 
2 & L17H24 & 75.52\% & L26H3 & 71.84\% \\ 
3 & MLP-29 & 74.45\% & MLP-29 & 71.37\% \\ 
4 & L31H1 & 74.38\% & L17H24 & 70.90\%  \\
5 & L30H27 & 71.04\% & MLP-28 & 70.03\%  \\
\end{tabular}
\caption{Top 5 performing components for Llama-3.1-8B-Instruct and Llama-3.1-8B-UltraMedical on JU task. L26H3 means the 3rd attention head at layer 26. MLP-29 means the MLP at layer 29.}
\label{tab:head_rank_JU}
\end{table}

\begin{table}[t]
\centering
\small
\setlength{\tabcolsep}{4pt}
\begin{tabular}{l r r r r} 
\textbf{Rank} & \textbf{Instruct} & \textbf{Accuracy} & \textbf{UltraMedical} & \textbf{Accuracy} \\ 
\midrule
1 & L18H20 & 61.34\% & L24H24 & 58.63\% \\ 
2 & L25H28 & 61.00\% & L14H6 & 58.46\% \\ 
3 & MLP-27 & 60.64\% & MLP-19 & 57.86\% \\ 
4 & L31H1 & 60.47\% & L24H26 & 57.39\%  \\
5 & L22H19 & 60.03\% & L21H17 & 57.26\%  \\
\end{tabular}
\caption{Top 5 performing components for Llama-3.1-8B-Instruct and Llama-3.1-8B-UltraMedical on JI task.}
\label{tab:head_rank_JI}
\end{table}

% \section{Weighted Kendall's $\tau$ detail}
% This appendix specifies the ranked correlation metrics used to compare per-component decomposition accuracies across the Instruct and UltraMedical models (Section~\ref{sec:comp-analysis}). We use the weighted Kendall's $\tau$~\cite{vigna2015weighted}, which places greater weight on agreement among the highest-ranked components.
\section{Weighted Kendall's $\tau$ for component-rank agreement}
\label{app:wkendall}

Here we specify the ranked correlation metric used in Section~\ref{sec:comp-analysis}. We use the weighted Kendall's $\tau_w$ of \citet{vigna2015weighted}, which places greater weight on agreement among the highest-ranked components.

\paragraph{Definition.}
The standard Kendall's $\tau$ counts concordant versus discordant pairs across two rankings $r$ and $s$ of $n$ items:
\begin{equation}
\tau = \frac{C - D}{\binom{n}{2}},
\end{equation}
where $C$ and $D$ are the numbers of concordant and discordant pairs. Each pair contributes equally regardless of where it falls in the ranking, which makes standard $\tau$ a poor choice when interest is concentrated on the top of the ranking. The weighted variant assigns each pair $(i, j)$ a weight that depends on the ranks of its members:
\begin{equation}
\tau_w = \frac{\sum_{i<j} w_{ij}\,\mathrm{sgn}\!\big((r_i - r_j)(s_i - s_j)\big)}{\sum_{i<j} w_{ij}},
\end{equation}
with $w_{ij} = w(r_i) + w(r_j) + w(s_i) + w(s_j)$ and hyperbolic weighting $w(r) = 1/(r+1)$, where ranks are taken in descending order of the underlying score (highest-scoring item has rank 0). Pairs involving high-ranked items therefore contribute substantially more to $\tau_w$ than pairs in the tail. This property makes $\tau_w$ well-suited to our setting, where the highest-accuracy components are of primary interest.

\paragraph{Implementation.}
Because weighted $\tau$ ranks its inputs in ascending order by default, we negate the accuracy vectors so that higher-accuracy components receive lower (top-of-list) ranks, ensuring the weighting emphasizes agreement among the strongest components. Aside from that, we use the Scipy implementation with default parameters to compute the metric.

\section{Ablation Procedure}
\label{app:ablation}
Following~\citet{NEURIPS2019_2c601ad9},  we implement ablation via per-component gate variables $\xi \in \{0, 1\}$ applied multiplicatively to each component's output before it is written to the residual stream. Setting $\xi =0$ zeroes the component's direct contribution to the forward pass without modifying any learned parameters. Gates are applied via forward-pass hooks and take effect at every token position. 

For the $i$'th attention head $h_i^{(l)}$ of layer $l$, recall that the multi-head attention output can be written as a sum over per-head contributions,
\begin{equation}
a^{(l)} = \sum_{i=1}^{n} \xi_i^{(l)} h_i^{(l)} W_{O,i}^{(l)},
\end{equation}
Where $W_{O,i}^{(l)}$ is the corresponding slice of the output projection matrix. Ablating head $i$ corresponds to setting $\xi_i^{(l)}=0$, so its contribution $h_i^{(l)} W_{O,i}^{(l)}$ vanishes from $a^{(l)}$. For an MLP block $m^{(l)}$, we analogously apply a gate $\xi_m^{(l)}=0$ to the block's output, removing its contribution to the residual stream at layer $l$. Both interventions are applied at every token position throughout the forward pass and modify only the activation tensors; no learned parameters are altered.

Unlike the reweighting procedure in Section~\ref{sec: comp-rw}, which scales each component's direct contribution to the output logits, ablation modifies the residual stream itself. The ablated component's effect on all downstream layers is therefore also removed, not only its direct contribution to the final logits.

For the random-ablation control in Figure~\ref{fig:ablate_cumulative}, we sample components uniformly without replacement from the set of all components excluding the top-20 ranked by individual decomposition accuracy, using a fixed seed. The sampled set is then ablated cumulatively in the same manner as the top-$k$ condition.

\section{Individual Component Ablation}
\label{app:ind-ablation}
To complement the cumulative ablation analysis in Section~\ref{sec:comp-analysis}, we also report results from ablating each of the top-5 components individually. Each component is removed in isolation using the procedure described in Appendix~\ref{app:ablation}, and the resulting accuracy change relative to the full model is recorded. As a control, we also report the largest accuracy drop observed across 10 individually ablated random components (sampled outside the top-20 highest accuracy components). The individual ablation results are shown in~\ref{tab:ind_ablation}.
\begin{table}[t]
\centering
\resizebox{\columnwidth}{!}{%
\begin{tabular}{l rr rr}
%\toprule
& \multicolumn{2}{c}{\textbf{Jargon Understanding (JU)}} & \multicolumn{2}{c}{\textbf{Jargon Identification (JI)}} \\
%\cmidrule(lr){2-3} \cmidrule(lr){4-5}
\addlinespace[0.1cm]
& Instruct & UltraMedical & Instruct & UltraMedical \\
\midrule
Full model & 76.05\% & 74.10\% & 58.80\% & 53.10\% \\
\addlinespace[0.1cm]
%\midrule
1 & L17H24 ($-6.49$) & L17H24 ($-7.29$) & MLP-19 ($-0.43$) & L26H3 ($-0.13$) \\
2 & L30H27 ($-0.67$) &  MLP-29 ($-0.67$) & L17H24 ($-0.1$) & L31H1 ($-0.03$) \\
3 & MLP-28 ($-0.33$) & L22H19 ($-0.67$) & L30H26 ($-0.07$) & MLP-29 ($-0.03$) \\
4 & L31H1 ($-0.20$)  & L31H1 ($-0.27$) & L26H22 ($-0.07$) & L31H3 ($+0.03$) \\
5 & MLP-26 ($-0.20$) & L30H27 ($-0.2$) & L25H8 ($-0.03$) & L30H27 ($+0.03$) \\
\midrule
Random (max drop) & $-0.20$ & $-0.4$ & $-0.23$ & $-1.1$ \\
%bottomrule
\end{tabular}%
}
\caption{Individual ablation results for the top-5 components on JU and JI. Values denote accuracy change (\%) relative to the full model when a single component is ablated. Random (max drop) is the largest accuracy drop observed across 10 individually ablated random components sampled outside the top-20 performing components.}
\label{tab:ind_ablation}
\end{table}
On JU, attention head L17H24 stands out as singularly important in both models: ablating it alone drops accuracy by 6.49\% in Instruct and 7.29\% in UltraMedical, far exceeding the change from any other top-5 component and well outside the range observed under random ablation. The larger drop in UltraMedical suggests that it relies more heavily on this particular component than Instruct does, rather than distributing jargon knowledge more broadly.
% medical fine-tuning increases the model's reliance on this particular component rather than distributing jargon knowledge more broadly. 
Beyond L17H24, the remaining top components produce modest drops of less than 0.7\% when ablated individually.

On JI, no single component exhibits a comparable individual impact. The largest individual drops are 0.43\% (MLP-19 in Instruct) and 0.13\% (L26H3 in UltraMedical), and several ablations even produce small positive changes within the range of noise. Notably, the maximum random ablation drop on UltraMedical ($-1.10$) actually exceeds the largest top-component drop, indicating that no individual component is disproportionately important for this task. This pattern is consistent with the flat per-component accuracy distribution observed in Section~\ref{sec:comp-analysis} and the shallow cumulative ablation curves in Figure~\ref{fig:ablate_cumulative}. JI relies on aggregated contributions from many components rather than a few specialized ones.

\section{Experiment setup for component reweighting}

\label{app:expsetupcomponent}

We optimize the component weights $w \in \mathbb{R}^{1056}$ (1,024 attention heads + 32 MLP) separately for each task and each model. Weights are trained for 1000 epochs with SGD (lr = 0.01, $\lambda$ = 0.01) with early stopping (patience = 3) on the training loss computed on the JI training set (1,200 examples) and evaluated on the 2,990-item JI test set. The same procedure is applied to JU using its 4,183-item training set and 1,495-item test set.

\section{Details for statistical tests}
\label{app: stat_detail}
\paragraph{McNemar's test.} We use a two-sided exact McNemar test for two paired comparisons on the JI test set: (1) Llama-3.1-8B-Instruct versus Llama-3.1-8B-UltraMedical before component reweighting, and (2) Llama-3.1-8B-UltraMedical before versus after component reweighting. In both comparisons, predictions are evaluated on the same test examples, allowing correctness outcomes to be paired. McNemar’s test considers only discordant examples, where one condition produces a correct prediction and the other produces an incorrect prediction. Let \(n_{01}\) denote the number of examples incorrect under the first condition but correct under the second, and \(n_{10}\) the number correct under the first condition but incorrect under the second. Under the null hypothesis, these two types of discordant outcomes are equally likely.
\paragraph{Bootstrap confidence intervals.} For weighted Kendall’s \(\tau_w\), we estimate 95\% confidence intervals using paired bootstrap resampling. We repeatedly sample model components with replacement while preserving the correspondence between the same component in Instruct and UltraMedical, and recompute \(\tau_w\) for each resampled set. We use 10,000 bootstrap replicates and report the 2.5th and 97.5th percentiles of the resulting bootstrap distribution as the 95\% confidence interval. This procedure provides an estimate of the uncertainty in the observed component-rank agreement between the two models.

\end{document}